\documentclass[sigconf]{acmart}

\AtBeginDocument{%
  }

\setcopyright{acmlicensed}
\copyrightyear{2024}
\acmYear{2024}
\acmDOI{XXXXXXX.XXXXXXX}
\acmConference[Conference acronym 'XX]{Make sure to enter the correct conference title from your rights confirmation emai}{June 03--05, 2024}{XX, XX}
\acmISBN{978-1-4503-XXXX-X/24/06}

\usepackage{multirow}
\usepackage{subfigure}
\usepackage{float}
\usepackage{balance}
\usepackage{xspace}
\usepackage{tcolorbox}
\usepackage{array}
\usepackage{graphicx}
\usepackage{booktabs}
\usepackage{soul}
\usepackage{CJKutf8}
\usepackage{enumitem}

\newcommand{\mmidr}{MMIDR\xspace}
\newcommand{\ourdata}{$\text{MR}_{llm}^{2}$\xspace}

\begin{document}

\title{\mmidr: Teaching Large Language Model to Interpret Multimodal Misinformation via Knowledge Distillation}

\author{Longzheng Wang$^*$}
\email{wanglongzheng@iie.ac.cn}
\affiliation{%
  \institution{Institute of Information Engineering, CAS}
  \institution{School of Cyber Security, UCAS}
  \country{China}
}

\author{Xiaohan Xu$^*$}
\email{xuxiaohan@iie.ac.cn}
\affiliation{%
  \institution{Institute of Information Engineering, CAS}
  \institution{School of Cyber Security, UCAS}
  \thanks{Equal Contribution}
  \country{China}
}

\author{Lei Zhang}
\email{zhanglei0510@iie.ac.cn}
\affiliation{%
  \institution{Institute of Information Engineering, CAS}
  \institution{School of Cyber Security, UCAS}
  \country{China}
}

\author{Jiarui Lu}
\email{lujiarui@iie.ac.cn}
\affiliation{%
  \institution{Institute of Information Engineering, CAS}
  \institution{School of Cyber Security, UCAS}
  \country{China}
}

\author{Yongxiu Xu}
\email{xuyongxiu@iie.ac.cn}
\affiliation{%
  \institution{Institute of Information Engineering, CAS}
  \country{China}
}

\author{Hongbo Xu}
\email{hbxu@iie.ac.cn}
\affiliation{%
  \institution{Institute of Information Engineering, CAS}
  \country{China}
}

\author{Minghao Tang}
\email{tangminghao@iie.ac.cn}
\affiliation{%
  \institution{Institute of Information Engineering, CAS}
  \country{China}
}

\author{Chuang Zhang}
\email{zhangchuang@iie.ac.cn}
\affiliation{%
  \institution{Institute of Information Engineering, CAS}
  \country{China}
}

\renewcommand{\shortauthors}{Longzheng Wang et al.}


\begin{abstract}
Automatic detection of multimodal misinformation has gained a widespread attention recently. However, the potential of powerful Large Language Models (LLMs) for multimodal misinformation detection remains underexplored. Besides, how to teach LLMs to interpret multimodal misinformation in cost-effective and accessible way is still an open question. To address that, we propose MMIDR, a novel framework for multimodal misinformation interpretation and distillation reasoning, aiming to teach LLMs in providing fluent and high-quality textual explanations for their decision-making process. To convert multimodal misinformation into an appropriate instruction-following format, we present a data augmentation perspective and pipeline. This pipeline consists of a visual information processing module and an evidence retrieval module. Subsequently, we prompt the proprietary LLMs with processed contents to extract rationales for interpreting the authenticity of multimodal misinformation. Furthermore, we design an efficient knowledge distillation approach to distill the capability of proprietary LLMs in explaining multimodal misinformation into open-source LLMs. To explore several research questions regarding the performance of LLMs in multimodal misinformation detection tasks, we construct an instruction-following multimodal misinformation dataset and conduct comprehensive experiments. The experimental findings reveal that our MMIDR exhibits sufficient detection performance and possesses the capacity to provide compelling rationales to support its assessments.
\end{abstract}



\maketitle

\section{Introduction}
In the context of the widespread adoption of social media platforms and online news outlets, individuals enjoy the freedom to share daily information, express their opinions, and convey their emotions. However, this freedom accelerates the generation and spread of various forms of misinformation (e.g., fake news, rumors)~\cite{zubiaga2018detection}. Existing misinformation detection research has primarily focused on textual data~\citep{yu2017convolutional,10.1145/3477495.3531744}. Nevertheless, most posts on these platforms are not confined to a specific modality. Detecting misinformation presented in different modalities poses a greater challenge, as it necessitates evaluating the credibility of each modality and their combinations \citep{abdali2022multi,cao2020exploring,DBLP:conf/ijcai/SharmaAADMFHSN022}. 

In spite of notable advancements in comprehending and representing misinformation \citep{hu2022deep}, existing models still face enormous challenges due to the intricate nature of the misinformation creation process. Misinformation creators demonstrate the ability to manipulate various aspects of information, employ diverse writing strategies, and operate with elusive underlying motives \citep{hu2023bad}. Leveraging evidence retrieved from the internet taps into a vast repository of world knowledge, enhancing the detection of meticulously crafted misinformation \citep{hu2023mr2}. Notably, the retrieved evidence could be multimodal, encompassing both textual and visual elements, thereby posing a challenging task for conventional small language models (SLMs).

Recent advancements in the field of Large Language Models (LLMs), such as GPT-3 \citep{brown2020language}, InstructGPT \citep{ouyang2022training}, and GPT-4 \citep{openai2023gpt4}, have demonstrated remarkable capabilities, such as the ability to follow instructions \citep{ouyang2022training}, perform knowledge-intensive tasks \citep{DBLP:conf/naacl/RubinHB22,shi2023replug}, and address societal challenges\citep{jiang2023raucg,roy2023probing}. In recent years, scholarly interest has been growing in investigating the application of LLMs in detecting misinformation~\citep{buchholz2023assessing,li2023self,pan2023fact,bang2023multitask,caramancion2023harnessing,chen2023combating}. These studies aim to explore and develop effective methods that leverage the language understanding and reasoning abilities of LLMs to enhance the accuracy and robustness of misinformation detection. However, the potential of LLMs in generating explanations for misinformation remains underexplored. Previous works on the explainability of misinformation detectors have primarily relied on extraction methods \citep{shu2019defend,lu2020gcan,DBLP:conf/acl/LiNK21,DBLP:conf/coling/00050CLLC22}. Nevertheless, LLMs exhibit the ability to generate fluent, natural language-based explanations for the given misinformation while predicting its authenticity. This capability is more human-friendly compared to earlier approaches \citep{chen2023combating}. Therefore, how to teach LLMs to interpret misinformation remains an open question.

It has been demonstrated that LLMs can be naturally extended to multimodalities, such as GPT-4~\citep{openai2023gpt4}, Gemini~\citep{team2023gemini}, and GPT-4V~\cite{2023GPT4VisionSC}. Multimodal Large Language Models (MLLMs) possess the ability to understand and process information across multiple modalities, surpassing the capabilities of traditional language-only LLMs. This broadened comprehension of information across multiple modalities enables MLLMs to closely simulate human perception, thereby expanding the potential for a diverse array of real-world applications. While proprietary LLMs and MLLMs exhibit remarkable capabilities, they still have certain limitations. One notable drawback is their limited accessibility and higher cost. In contrast, open-source LLMs, exemplified by LLaMA~\citep{touvron2023llama}, offer distinct advantages. Additionally, recent studies have widely adopted knowledge distillation techniques, confirming their effectiveness in bridging the performance gap between proprietary and open-source LLMs \citep{hsieh2023distilling,yang2023gpt4tools,jung2023impossible,ramnath2023tailoring,liu2023minds,xu2024survey}. To the best of our knowledge, the research on utilizing LLMs and MLLMs for multimodal misinformation detection via knowledge distillation techniques remains underexplored.

Taking the considerations above, we propose \mmidr, a novel framework for \textbf{M}ultimodal \textbf{M}isinformation \textbf{I}nterpretation and \textbf{D}isti- llation \textbf{R}easoning. We present a data augmentation perspective and pipeline to convert image-text pairs into an appropriate instruction-following format. Specifically, for a given multimodal misinformation, we utilize OCR and image captioning techniques to process visual information. Additionally, we obtain both visual and textual evidence through evidence retrieval \citep{hu2023mr2}. After that, we feed the processed contents into proprietary LLMs (e.g., ChatGPT~\citep{ouyang2022training}) and extract relevant rationales aimed at explaining the authenticity of multimodal misinformation. Finally, we design an efficient knowledge distillation approach. By integrating the original multimodal information content with the rationales provided by proprietary LLMs, also known as teacher LLMs, we employed the LoRA (Low-Rank Adaptation) \citep{hu2021lora} technology to fine-tune open-source LLMs (e.g., LLaMA~\citep{touvron2023llama}, MiniGPT-v2~\citep{chen2023minigpt}), also known as student LLMs. In this way, \mmidr can distill the capability of teacher LLMs in explaining multimodal misinformation into student LLMs. This process can promote a comprehensive understanding and integration of multimodal misinformation content by open-source LLMs, empowering them to generate high-quality rationales to interpret their decision-making processes.

The main contributions of this paper are as follows:
\begin{itemize}
    \item We investigate the utilization of LLMs and MLLMs for the detection of multimodal misinformation. To our best knowledge, this is the first exploration in this area.
    \item We propose \mmidr, a framework designed to teach LLMs to interpret multimodal misinformation, aiming to provide fluent and high-quality textual explanations for their decision-making process.
    \item We introduce a data augmentation perspective and pipeline that converts multimodal retrieval-enhanced misinformation data into an appropriate instruction-following format. This process enables the acquisition of knowledge from proprietary LLMs, leading to the construction of a dataset \ourdata.
    \item We propose an efficient knowledge distillation approach designed to address the limited accessibility and higher cost of proprietary LLMs. Simultaneously, our approach aims to bridge the performance gap between open-source LLMs and proprietary LLMs, empowering the former to generate high-quality explanations.
    \item We conduct extensive experiments on the dataset \ourdata, investigating several research questions regarding the performance of LLMs in multimodal misinformation detection tasks. Experimental results demonstrate that our framework has sufficient detection performance and can provide compelling rationales to support its assessments.
\end{itemize}


\section{Related Works}
\subsection{Multimodal Misinformation Detection}
Existing misinformation Detection methods~\citep{qian2018neural,bhattarai2021explainable,guo2018rumor,potthast2017stylometric,conroy2015automatic}mostly fall into text-based. Some previous works have explored the explainability of misinformation detectors. dEFEND \citep{shu2019defend} leverages a combined analysis of news content and user comments to capture explainable user comments for misinformation detection. GCAN \citep{lu2020gcan} employs a dual co-attention mechanism to better learn the representations of user interactions, retweet propagation, and their correlation, thereby enabling the generation of reasonable explanations. However, previous works on the explainability of misinformation detectors have primarily relied on extraction methods, which may lack human-friendliness. In this paper, we effectively leverage LLMs to generate fluent, natural language-based explanations for interpreting misinformation




More recently, several methods \citep{wang2018eann,khattar2019mvae,wu2021multimodal,chen2022cross,singhal2022leveraging,sun2023inconsistent,wang2023cross} have been proposed to evaluate the credibility of each modality and their combinations, achieving superior performance in multimodal misinformation detection. For instance, EANN~\citep{wang2018eann} employs an adversarial approach to remove event-specific features, thereby enhancing the adaptability to newly emerging events. KDCN~\citep{sun2023inconsistent} suggests a unified framework for detecting multimodal misinformation by modeling cross-modal and content-knowledge inconsistencies.  CAFE~\cite{chen2022cross} measures cross-modal ambiguity through the assessment of Kullback-Leibler (KL) divergence between unimodal feature distributions. COOLANT~\citep{wang2023cross} leverages cross-modal contrastive learning to achieve more accurate image-text alignment.  However, these approaches focus solely on the multimodal information without exploiting the internet to retrieve evidence, which could enhance detection performance \citep{hu2023mr2}. Given that the retrieved evidence may consist of both textual and visual components, conventional small language models face challenges in processing such multimodal data. In this study, we advocate the utilization of large language models to tackle these intricate scenarios.

\subsection{LLMs for Detecting Misinformation}
In recent years, there have been increasing efforts to investigate the application of LLMs in detecting misinformation~\citep{buchholz2023assessing,li2023self,pan2023fact,bang2023multitask,caramancion2023harnessing,chen2023combating,pelrine2023towards,pavlyshenko2023analysis,cheung2023factllama,pan2023fact, zhang2023towards}. These studies can be divided into three methods: 1) directly prompting-based methods, 2) fine-tuning-based methods, and 3) augmentation-based methods, which augment LLMs with external knowledge or tools. For instance, ARG \cite{hu2023bad} employs a small model to encode specific news along with the textual and common-sense descriptions generated by LLMs. DELL \cite{wan2024dell} utilizes LLMs to generate synthesized reactions and comments from diverse viewpoints, employing them for proxy tasks related to prediction and explanation. \cite{pavlyshenko2023analysis} utilize LoRA to fine-tune LLaMA2 \citep{touvron2023llama} aiming to adapt it for various tasks. FACTOOL \cite{chern2023factool} proposes a multi-task and multi-scenario framework aimed at detecting factual errors in LLM-generated texts. While LLMs can naturally expand to multimodalities, the extensive potential they hold for combating multimodal misinformation in real-world settings remains largely unexplored.

\subsection{Instruction Tuning}
Recent studies \citep{ouyang2022training,wang2022self,wang2022benchmarking}  have explored methods for refining the instruction-following capabilities of LLMs. These studies reveal that by fine-tuning on specific instruction-following datasets, LLMs can follow natural language instructions, enabling them to perform various real-world tasks effectively. More importantly, there has been a surge in the adoption of knowledge distillation techniques \citep{hsieh2023distilling,yang2023gpt4tools,jung2023impossible,ramnath2023tailoring,liu2023minds,gou2021knowledge,gupta2022compression,xu2024survey}, focusing on distilling off-the-shelf language models using instructional data obtained from strong LLMs. This trend aims to enhance the zero- and few-shot abilities of language models. For instance, the Evol-Instruct method \citep{xu2023wizardlm} expands the instructions in two dimensions: difficulty, such as the rephrasing of questions to heighten complexity, and diversity, entailing the generation of long-tailed instructions. This method has been used to expand the distillation of coding \citep{luo2023wizardcoder} and math \citep{luo2023wizardmath}. In the vision-language domain, LLaVA \citep{liu2024visual} translates images into textual descriptions, including captions and bounding boxes, and subsequently employs GPT-3.5 and GPT-4 for distillation to generate instruction-following data related to visual content. In this paper, we utilize GPT-3.5 to extract rationales and construct an instruction-following multimodal misinformation dataset. We employ LoRA to train student LLMs (e.g., LLaMA, MiniGPT-v2), enabling them to interpret
multimodal misinformation effectively.

\begin{figure*}[t]
  \centering
  \includegraphics[width=0.95\linewidth]{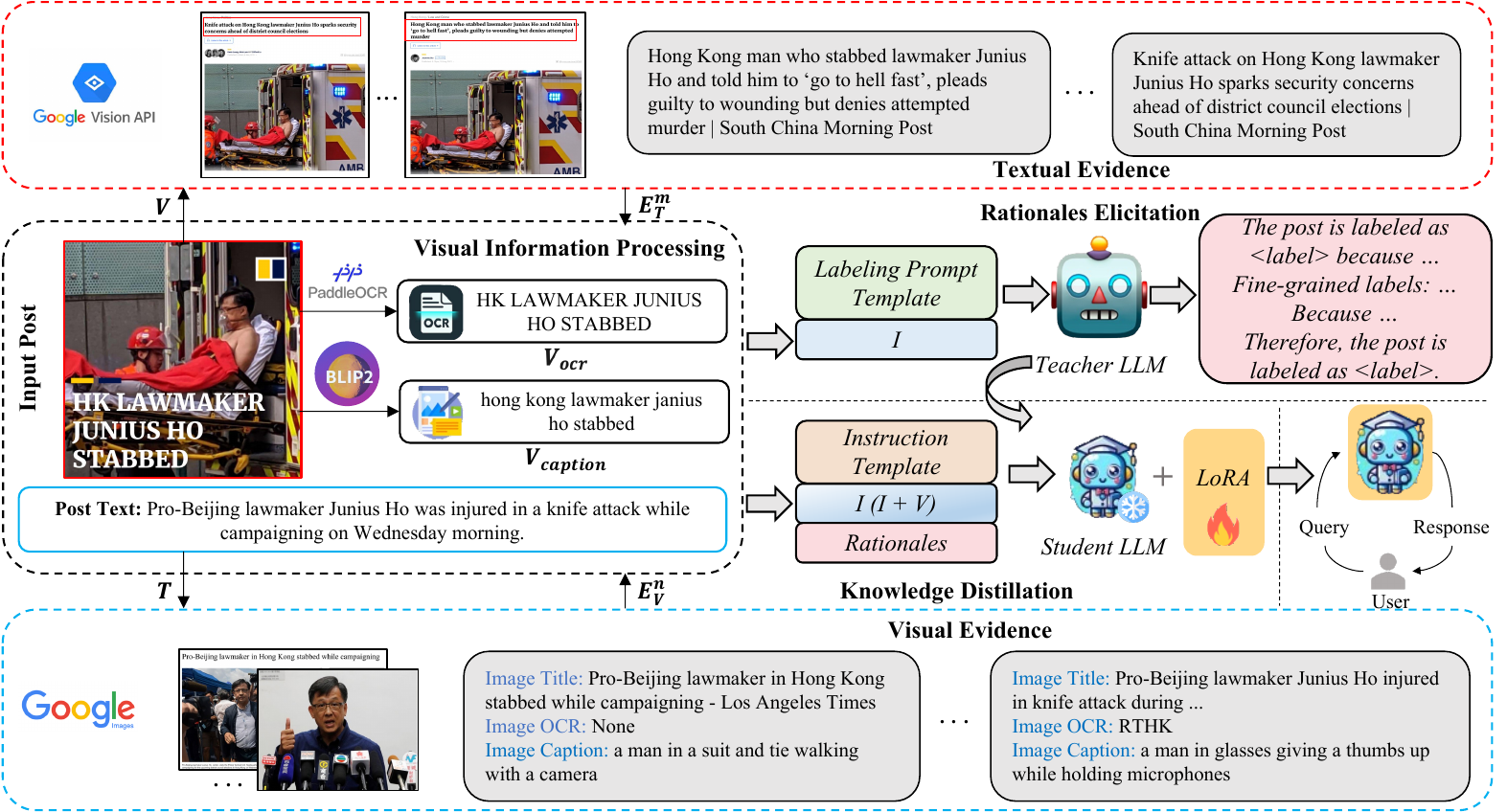}
  \caption{Model Architecture Overview of \mmidr. Given multimodal information, we first convert it into an appropriate instruction-following format. Then we employ the labeling template to prompt the teacher LLM in order to obtain rationales for the authenticity labels assigned to provided multimodal information. Finally, we employ LoRA~\citep{hu2021lora} to train the student LLM on the integration of the original multimodal information and the rationales.
  $I(I+V)$ denotes that the input for student LLM could be textual instruction ($I$) or both textual instruction and image ($I+V$) based on whether student is multimodal.}
  \label{fig:mr2_model}
\end{figure*}

\section{Methodology}~\label{method}
In this section, we introduce our novel framework \mmidr, designed specifically to teach LLMs to interpret multimodal misinformation. The primary objective of \mmidr is to provide fluent and high-quality textual explanations for the decision-making process. The overall model structure is illustrated in Figure~\ref{fig:mr2_model}. For each given multimodal information, we first conduct data augmentation (\S\ref{sec:augmentation}), which involves transforming the multimodal retrieval-enhanced misinformation into an appropriate instruction-following format. Subsequently, we input this processed content into teacher LLMs for the extraction of rationales (\S\ref{sec:extracting}). Finally, we efficiently perform knowledge distillation to distill the capability of teacher LLMs in explaining multimodal misinformation into student LLMs (\S\ref{sec:finetune}). 

\subsection{Data Augmentation} \label{sec:augmentation}
For each input multimodal information $\mathbf{P} = \{T, V\} \in \mathcal{D}$, where $T$, $V$, and $\mathcal{D}$ represent text, image, and the original dataset, respectively. This work requires the application of large language models to guide and extract rationales (\S\ref{sec:extracting}). To facilitate this, we convert the original multimodal information's image $V$ into textual information compatible with the large language model (\S\ref{sec:textgen}). Subsequently, we use text $T$ and image $V$ separately for retrieval purposes, aimed to procure evidence that supports the large language model's reasoning process (\S\ref{sec:evidence}).

\subsubsection{Visual Information Processing} \label{sec:textgen}
Given visual content $V$, we leverage Optical Character Recognition (OCR) technology to discern and transmute characters embedded within the image into an editable text format, designated as $V_{ocr}$. This process is specifically executed through the utilization of Baidu's PaddleOCR framework\footnote{https://github.com/PaddlePaddle/PaddleOCR}.Simultaneously, we employ Image Captioning technology to provide descriptions for the image content. In particular, we choose to employ the BLIP-2 $\text{OPT}_{2.7B}$ model\citep{li2023blip} for generating descriptive text, resulting in $V_{caption}$.

\subsubsection{Evidence Retrieval} \label{sec:evidence}
Through the retrieval of evidence from the internet, the model is empowered to embrace a broader scope of global knowledge, thereby facilitating the identification of meticulously crafted misinformation. Drawing inspiration from the evidence retrieval methodology\citep{hu2023mr2}, our approach involves utilizing the image $V$ from the input post as a query. By employing the Google Reverse Image\footnote{https://cloud.google.com/vision}, we systematically retrieve textual evidence denoted as $E_T$. The search engine yields a selection of images resembling the query image, from which we extract textual evidence, including titles and descriptions.

Similarly, we employ the textual content $T$ from the input post as a query and leverage the Google Programmable Search Engine\footnote{https://programmablesearchengine.google.com} to obtain visual evidence denoted as $E_V$. The search engine generates a collection of images corresponding to the query text. For these visual evidentiary elements, we implement the aforementioned visual information processing techniques ($\S$\ref{sec:textgen}), encompassing OCR extraction and image caption generation. Hence, a piece of visual evidence $E_V = \{Image\_Title, Image\_OCR, Image\_Caption\}$ represents the corresponding image's title, OCR extraction results, and image caption generation results, respectively.

\subsection{Rationales Elicitation} \label{sec:extracting}
Rationales elicitation refers to using a teacher LLM to generate explanations for the given multimodal information. Specifically, each instance is represented as a tuple $x = \{T, V_{ocr}, V_{caption}, E^{m}_T,  E^{n}_V \}$, comprising the textual content of the post $T$, OCR-recognized text derived from visual content $V_{ocr}$, text generated from image captions associated with the visual content $V_{caption}$, along with a collection of textual evidence $E^{m}_T$ and visual evidence $E^{n}_V$, where $m$ and $n$ denote the number of pieces of evidence.

To utilize the instruction-following ability of LLMs~\citep{wei2021finetuned}, we design a simple labeling prompt template $I$ that prompts the teacher LLM to generate rationales:

\noindent
\begin{minipage}[t]{1\columnwidth}
    \label{fg:template}
    \centering
    \resizebox{1\columnwidth}{!}{
    \begin{tcolorbox}[
    title=Labeling Prompt Template, 
    ]
    \small
    Now give you a multimodal post and the corresponding textual evidences and image evidences, these evidences may be relevant or irrelevant.
    
    Each post is a tuple of (text, image\_ocr, image\_caption). 
    
    Evidences are a list of sentences, which are split by "<and>".

    \#\# Three-way classification scheme
    
    Each post is labeled in three-way classification scheme: [0: "non-rumor", 1: "rumor", 2: "unverified"].

    \#\# Fine-grained labels
    
    \colorbox[HTML]{e5f1db}{<Non-rumor | Rumor | Unverified fine-grained labels>}

    \#\# Post
    
    Post:[
    
        \quad text: \colorbox[HTML]{e0ebf6}{\{post\_text\}}, 
        
        \quad image ocr: \colorbox[HTML]{e0ebf6}{\{image\_ocr\}}, image caption: \colorbox[HTML]{e0ebf6}{\{image\_caption\}}]

    \#\# Evidences
    
    Textual\_evidence: \colorbox[HTML]{e0ebf6}{\{textual\_evidence\}}
    
    Image\_evidence: \colorbox[HTML]{e0ebf6}{\{visual\_evidence\}}
    
    \#\# Label
    
    label: \colorbox[HTML]{e5f1db}{<Non-rumor | Rumor | Unverified>}

    \#\# Explanation
    
    Now, you are required to explain why the post is labeled as \colorbox[HTML]{e5f1db}{<Non-rumor | Rumor | Unverified>} according to the evidences, and decide which fine-grained labels are suitable for the post.
    \end{tcolorbox}
    }
\end{minipage}

To instantiate the template, each instance $x$ is incorporated into the designated position within the template's blue segment. Simultaneously, based on its ground-truth label $g$, the template's green segment is specifically populated. The comprehensive details regarding fine-grained labels \citep{hu2023mr2} can be found in Table \ref{tb:fine-grained}. An instantiated template is denoted as $I(x,g)$.  

By feeding the instantiated template $I(x,g)$ into the teacher LLM, we could annotate the provided input $x$ through the labeling prompt template template, yielding the output $y$ of the rationale. This process can be formulated as follows:
\begin{equation}
    \mathcal{D}^{\text{(rationales)}} = \{x, y|x \sim \mathcal{D}, y \sim p_{T}\left(y |I(x,g)\right)\}.
\end{equation}
where $I(x,g)$ is a template instantiated by the given instance $x$ and the ground-truth label $g$. $p_{T}$ represents the teacher LLM with parameters $\theta_T$. Additionally, to guarantee the output is ended with the prediction of the label, each output $y$ is appended with a specific text of \textit{"Therefore, the post is labeled as <label>"}. Therefore, the output $y$ encompasses both the rationale and the ground-truth label, which could be used to train the student LLM subsequently. 

\begin{table*}[!ht]
\small
\centering
\caption{Mappings from fine-grained labels to three standardized labels.}
\resizebox{0.99\textwidth}{!}{
\begin{tabular}{cc}
\toprule
\bf Standardized & \bf Fine-grained Labels \\
\midrule
\multirow{7}{*}{\textbf{Rumor}} 
& Altered, Altered Image, Altered Photo, Bad Math, Bad Science, Clickbait, Commontion, Death Hoax, Distorts the Facts, Doctored,  \\
& FAKE, Fake News, Fake Quotes, Fake Quoto, Fake Tweet, False, False and Misleading, False Headline, False!, Flawed-Reasoning, Flip-flop, \\
& Full Flop, Hoax, Hoax!, Inaccurate, Incorrect, It’s A Joke, Lacks Context, Likely False, Misattributed, Misleading, Misleading and False, \\
& Misleading!, Misleading., Misplaced Context, Misrepresented, Missing Context, Mixed, Mixture, Mostly False, Needs Context, Not Legit, \\
& Not the full story, Not the Whole Story, Out of Context, Pants on Fire, Partly False, Photoshopped, Selective, Spins the Facts, Staged Skit, \\
& Suspicious, This claim is False., This is misleading., This is not true., Three Pinocchios, Totally Fake, Totally False, Trolling, \\
& Two Pinocchios, Wrong, Wrong Number, Wrong Numbers \\
\midrule
\multirow{1}{*}{\textbf{Non-Rumor}} 
& Accurate, Close to the mark, Correct, Mostly correct, Mostly True, Mostly-Accurate, No Fraud, Partially True, Partially-Correct, True \\
\midrule
\multirow{1}{*}{\textbf{Unverified}} 
& Baseless, Myth, No Evidence, No Proof, Not Supported, This lacks evidence., Unclear, Unproven, Unsubstantiated, Unsupported \\
\bottomrule
  \end{tabular}}
  \label{tb:fine-grained}
\end{table*}

\subsection{Knowledge Distillation} \label{sec:finetune}
Knowledge distillation refers to fine-tune the student LLM by maximizing the likelihood of sequences generated by the teacher LLM, ensuring alignment between the predictions of the student and the teacher. Based on the dataset $\mathcal{D}^{\text{(rationales)}}$, we tune the off-the-shelf language model, adhering to its original auto-regressive training objective. To facilitate this tuning process, we employ LoRA~\citep{hu2021lora} optimization, wherein the language model is frozen, and optimization is solely applied to the rank decomposition components of the Transformer layers.

Due to the absence of label during inference, we design a new instruction template $I'$ derived from the labeling prompt template $I$ by removing label-related contents, i.e. the green segments in the template. That is, instantiating the $I'$ only requires the instance $x$, i.e. $I'(x)$. For an output $y$ comprising $L$ tokens, the learning objective of student LLM is to maximize the conditional language modeling:
\begin{equation}
    p\left(y \mid I'(x) \right)=\prod_{i=1}^{L} p_{\theta}\left(y_{i} \mid I'(x), y_{1: i-1}\right)
\end{equation}
where $I'(x)$ denotes the instantiated instruction, and $\theta$ represents the trainable parameters. Considering that the student LLM adopted in our study encompasses multimodal large language models, its input component is equipped to accommodate the original visual content. Therefore, the aforementioned process can be described as follows:
\begin{equation}
    p\left(y \mid V, I'(x)\right)=\prod_{i=1}^{L} p_{\theta}\left(y_{i} \mid V,I'(x), y_{1: i-1}\right)
\end{equation}
where $V$ is the original visual content.

\section{Experiments}
In our experimental endeavors to explore the utilization of LLMs in the realm of multimodal misinformation detection, we aim to address the following research questions:
\begin{itemize}
    \item RQ1: Do closed-source LLMs and open-source LLMs possess sufficient background knowledge to detect multimodal misinformation?
    \item RQ2: Can fine-tuned open-source LLMs independently serve as multimodal misinformation detectors?
    \item RQ3: After knowledge distillation, do open-source LLMs demonstrate sufficient detection performance and provide compelling rationales to support their assessments?
\end{itemize}

\subsection{Experimental Configurations}
\subsubsection{Datasets}
The experimental dataset is constructed based on the original dataset $\text{MR}^{2}$~\citep{hu2023mr2}. The dataset comprises two modalities, encompassing both images and text, and incorporates two languages, English and Chinese. By leveraging the remarkable cross-lingual capacities inherent in LLMs, the dataset developed in our study no longer strictly segregates between languages.

\begin{figure}[htbp]
  \centering
  \includegraphics[width=1\linewidth]{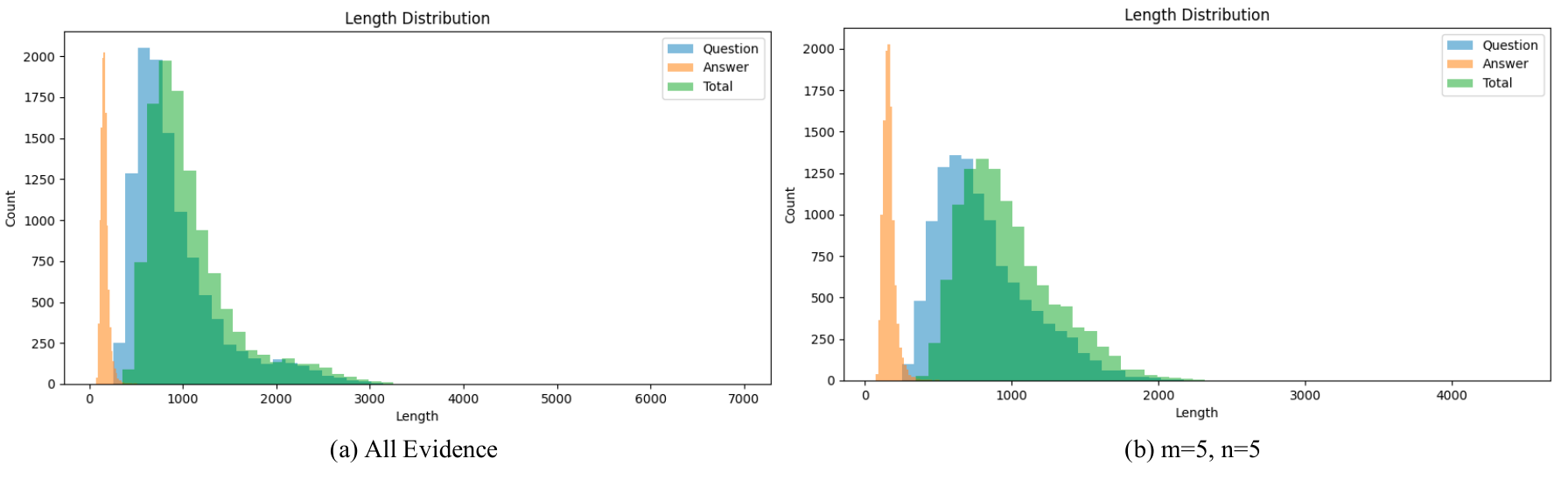}
  \caption{The length distribution of training data composed of different numbers of evidence.}
  \label{fig:evidence differentiation}
\end{figure}

We employ ChatGPT (gpt-3.5-turbo) \citep{ouyang2022training} as the teacher model to elucidate the rationales behind labeling multimodal misinformation based on evidence and determining fine-grained labels applicable to each post. The result instruction-following dataset is denoted as \ourdata. The statistical features and labels distribution of \ourdata are presented in Table \ref{table: our mr2 dataset}. Additionally, we graphically depict the length of instances and generated output with varying quantities of textual evidence ($m$) and visual evidence ($n$) obtained through the evidence retrieval module ($\S$\ref{sec:evidence}), as illustrated in Figure \ref{fig:evidence differentiation}. Specifically, we instantiate instances with different quantities of evidence using the instruction template, concatenating the rationales generated by the teacher model for illustration. We provide visualizations of the lengths of all instances, which serve as a basis for subsequent experiments and as references in hyperparameter settings.

\begin{table}[t]
\centering
\caption{Statistics of \ourdata dataset and the labels distribution.}
\setlength{\tabcolsep}{1mm}{ 
	\begin{tabular}{ccccc}
		\toprule
		Set & Non\_rumor & Rumor &  Unverified  &  Total \\
		\midrule
		Train & 3,592 & 2,392 & 5,200 & 11,184\\
		Test  & 421 & 280 & 608 &1,309 \\
		\midrule
		Total &4,013 &2,672 &5,808 &12,493 \\
		\bottomrule
\end{tabular}}
	\label{table: our mr2 dataset}
\end{table}

\subsubsection{Evaluation Metrics and Baselines}
The evaluation metrics include Accuracy, Precision, Recall, and F1-score. We compare \mmidr with 3 baseline model for multimodal misinformation detectio: \textbf{1)} BERT~\citep{devlin2018bert}+ ResNet~\citep{he2016deep}, \textbf{2)} CAFE~\citep{chen2022cross}, leveraging cross-modal ambiguity measurement for adaptive aggregation, and \textbf{3)} COOLANT~\citep{wang2023cross}, employing cross-modal contrastive learning to enhance image-text alignment accuracy.


\subsubsection{Implementation Details}
For \mmidr, we use LLaMA2-Chat-7B \citep{touvron2023llama} as the LLM backbone, which was pretrained on 2 trillion tokens of data from publicly available sources. We employ MiniGPT-v2 \citep{chen2023minigpt} as the foundational architecture for our MLLM. MiniGPT-v2 directly assimilates visual tokens from a ViT vision encoder \citep{Fang_2023_CVPR} and projects them into the feature space of an LLM \citep{touvron2023llama}.

Based on the constructed data, we perform fine-tuning of language models employing LoRA technology \citep{hu2021lora}. Specifically, we equip the projection layers of query, value, and output with LoRA layers. The LoRA attention dimension and scaling alpha are set to 16. While maintaining the language model in a frozen state, the LoRA layers are optimized using the AdamW \citep{loshchilov2018decoupled}. All models are fine-tuned over 50 epochs, with a batch size 64. The learning rate is set to $1 \times 10^{-4}$, and the maximum length of newly generated tokens is constrained to 2048.

\subsection{RQ1: Zero-shot Experiments}
In this section, we perform zero-shot experiments to verify whether large language models inherently possess enough background knowledge to detect multimodal misinformation. Specifically, we evaluate both ChatGPT (gpt-3.5-turbo) \citep{ouyang2022training} and MiniGPT-v2 \citep{chen2023minigpt} on the test set of \ourdata. The results are presented in Table \ref{tab:zero_shot}.

From the results, we can notice that the performance of all LLMs does not attain a satisfactory level\footnote{Detailed comparisons are not conducted here due to the limited compatibility of LLaMA2-7B-Chat~\citep{touvron2023llama} with instructions.}. With evidence support, the accuracy of the teacher model gpt-3.5-turbo is below 50\%. The classification performance of MiniGPT-v2 is slightly better than that of gpt-3.5-turbo, suggesting that the introduction of original visual content contributes to the detection of multimodal misinformation. It is evident that even powerful LLMs may lack sufficient background knowledge to detect multimodal misinformation.

\begin{table}[t]
\centering
\caption{Zero-shot Experiments with gpt-3.5-turbo and MiniGPT-v2.}
\setlength{\tabcolsep}{1.6mm}{ 
	\begin{tabular}{ccccc}
		\toprule
		Models  & Accuracy & Precision   & Recall  & F1  \\
		\midrule
		  gpt-3.5-turbo & 48.89 & 42.91 & 42.04 & 41.56 \\
	- w/o textual evidence & 49.89 & 42.13 & 42.25 & 41.18 \\
	- w/o visual evidence & 49.73 & 42.98 & 40.42 & 38.30 \\
        - w/o any evidence & 36.75 & 38.54 & 32.48 & 32.94 \\
		\midrule
            MiniGPT-v2 & \bf 53.72 & \bf 53.70 & \bf 53.38 & \bf 52.52 \\
            \bottomrule
\end{tabular}}
	\label{tab:zero_shot}
\end{table}

\begin{figure}[htbp]
  \centering
  \includegraphics[width=0.82\linewidth]{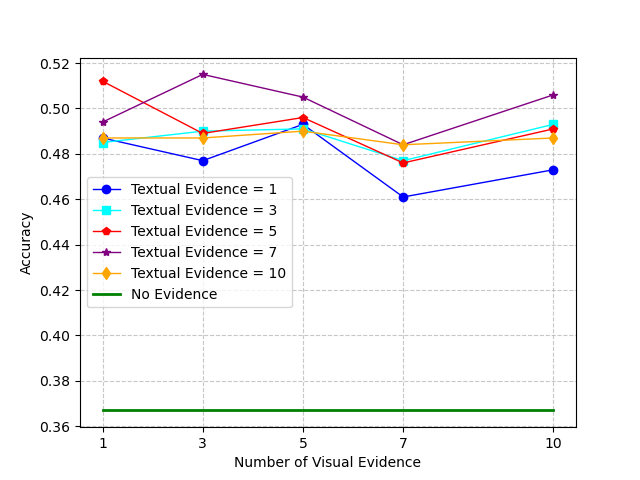}
  \caption{The classification accuracy of the teacher model using different numbers of evidence.}
  \label{fg:evidence_count}
\end{figure}

Additionally, it is noted that furnishing either solely textual evidence or visual evidence yields a marginally higher accuracy for the teacher model, compared to the simultaneous presentation of all evidence. This phenomenon may be attributed to the potential challenge posed to the teacher model in processing an excessive volume of evidence. Therefore,  we initiated a comprehensive examination to discern the influence of evidence quantity on the classification accuracy of the teacher model. In Figure~\ref{fg:evidence_count}, we vary the numbers of retrieved visual and textual evidence from $1 \sim 10$ and report the Accuracy on the test set. It is evident that the quantity of evidence significantly impacts the decisions made by LLMs, reaching an optimal effect when the cumulative quantity of visual and textual evidence is 6, as depicted by the red and blue lines. Nevertheless, these effects do not surpass the performance achieved by MiniGPT-v2 (53.72\%). Furthermore, the fluctuation results further indicate the challenges faced by LLMs in effectively detecting multimodal misinformation.

\subsection{RQ2: Performance Comparison and Ablation Study}
In this section, we compare \mmidr with other baselines on multimodal misinformation detection tasks. Note that $\text{\mmidr}_{\text{LLaMA2}}$ and $\text{\mmidr}_{\text{MiniGPT-v2}}$ denote the results obtained by fine-tuning \mmidr model on LLaMA2~\citep{touvron2023llama} and MiniGPT-v2 \citep{chen2023minigpt}, respectively. We also compare two variants of \mmidr: \textbf{1)} \textit{w/o evidence\&rationale}, where we remove evidence from the instruction during the detection of multimodal misinformation. Given that rationales encompass descriptions of certain evidence, they should be excluded in this context. \textbf{2)} \textit{ w/o rationale}, where we simply eliminate the rationales. Hence, these two variants can be considered equivalent to inputting multimodal misinformation and having the LLMs directly generate the veracity label. The results are shown in Table~\ref{tab:main_comp}. From the results, we observe that:
\begin{itemize}
    \item The fine-tuned baseline models show a significant improvement in accuracy compared to the zero-shot experiments of LLMs, demonstrating that smaller models can achieve higher classification performance after task-specific training.
    \item After being fine-tuned with \mmidr \textit{w/o rationale}, both LLaM-A2 and MiniGPT-v2 achieve optimal performance, reaching 95.83\% and 95.95\%, respectively. Compared to the baseline model BERT+ResNet, there is an improvement of over 20\%. This demonstrates that LLMs have the capability to handle multimodal misinformation detection tasks. Furthermore, the inherent robust background knowledge of LLMs enables them to adapt effectively to various tasks with minimal information.
    \item After being fine-tuned with \mmidr \textit{w/o evidence\&rationale}, the performance is slightly worse compared to \mmidr \textit{w/o rationale} (95.57\% vs. 95.83\% \& 94.19\% vs. 95.95\%), further confirming that evidence has a certain effect on enhancing the model's detection performance.
    \item In the main experimental section of \mmidr, there is a slight performance decline compared to the two aforementioned variants. We attribute this to two factors: \textbf{1)} The two aforementioned variants exclusively employ LLMs for a relatively straightforward three-classification task; and \textbf{2)} The objective function during the distillation process primarily emphasizes the alignment of predictions between the student model and the teacher model, omitting the inclusion of task-specific classification loss. Nevertheless, our primary goal is the distillation of open-source LLMs/MLLMs through the utilization of rationales generated by the teacher model, facilitating the production of high-quality, cohesive textual explanations for their decision rationale. In pursuit of this objective, a marginal performance decrease is deemed acceptable (93.63\% vs. 95.83\% \& 94.04\% vs. 95.95\%).
\end{itemize}

\begin{table}[t]
\centering
\caption{Performance comparison between \mmidr and other methods on \ourdata dataset. The best two results are \textbf{bolded} and \underline{underlined}, respectively. For the main experimental results of the \mmidr, apply \textcolor{red}{red} font for annotation.}
\setlength{\tabcolsep}{1.2mm}{ 
	\begin{tabular}{ccccc}
		\toprule
		Models  & Accuracy & Precision   & Recall  & F1  \\
            \midrule
		  gpt-3.5-turbo(zero-shot) & 48.89 & 42.91 & 42.04 & 41.56 \\
            MiniGPT-v2(zero-shot) &  53.72 & 53.70 & 53.38 &  52.52 \\
		\midrule
		  BERT+ResNet & 75.40 & 75.34 & 72.51 & 73.61 \\
            CAFE & 78.69 & 78.28 & 74.39 & 75.71 \\
            COOLANT & 82.97 & 84.88 & 77.49 & 79.56 \\
            \midrule
            $\text{\mmidr}_{\text{LLaMA2}}$ & \textcolor{red}{93.63} & \textcolor{red}{93.62} & \textcolor{red}{93.62} & \textcolor{red}{93.62} \\
            - w/o evidence\&rationale& 95.57 & 95.50 & 95.66 & 95.56 \\
            - w/o rationale & \underline{95.83} & \underline{95.80} & \underline{95.87} & \underline{95.83} \\
            \midrule
            $\text{\mmidr}_{\text{MiniGPT-v2}}$ & \textcolor{red}{94.04} & \textcolor{red}{94.12} & \textcolor{red}{93.90} & \textcolor{red}{94.00} \\
            - w/o evidence\&rationale& 94.19 & 94.22 & 94.11 & 94.16 \\
            - w/o rationale & \bf 95.95 & \bf 95.97 & \bf 95.89 & \bf 95.92 \\
            \bottomrule
\end{tabular}}
	\label{tab:main_comp}
\end{table}

\begin{table*}[t]
\small
\begin{minipage}{0.99\textwidth}
\centering
\caption{Examples from \ourdata (Chinese is translated
into English). We annotate instructions for the LLMs, highlighting data augmentation content in {\color{blue} blue}, which includes OCR, captions, and evidence. Key evidence is indicated in {\color{brown} brown}. In the generated explanations, we mark in {\color{purple} purple} the parts that most clearly indicate misinformation.}
\resizebox{!}{0.44\linewidth}{
\begin{tabular}{l >{\centering\arraybackslash}p{10cm} >{\centering\arraybackslash}p{9.6cm}}
\toprule
\multicolumn{3}{l}{\bf Examples from \ourdata:}  \\
\midrule
\multirow{2}{*}{\bf Input Post} &  \includegraphics[width=3.5cm, height=3.5cm]{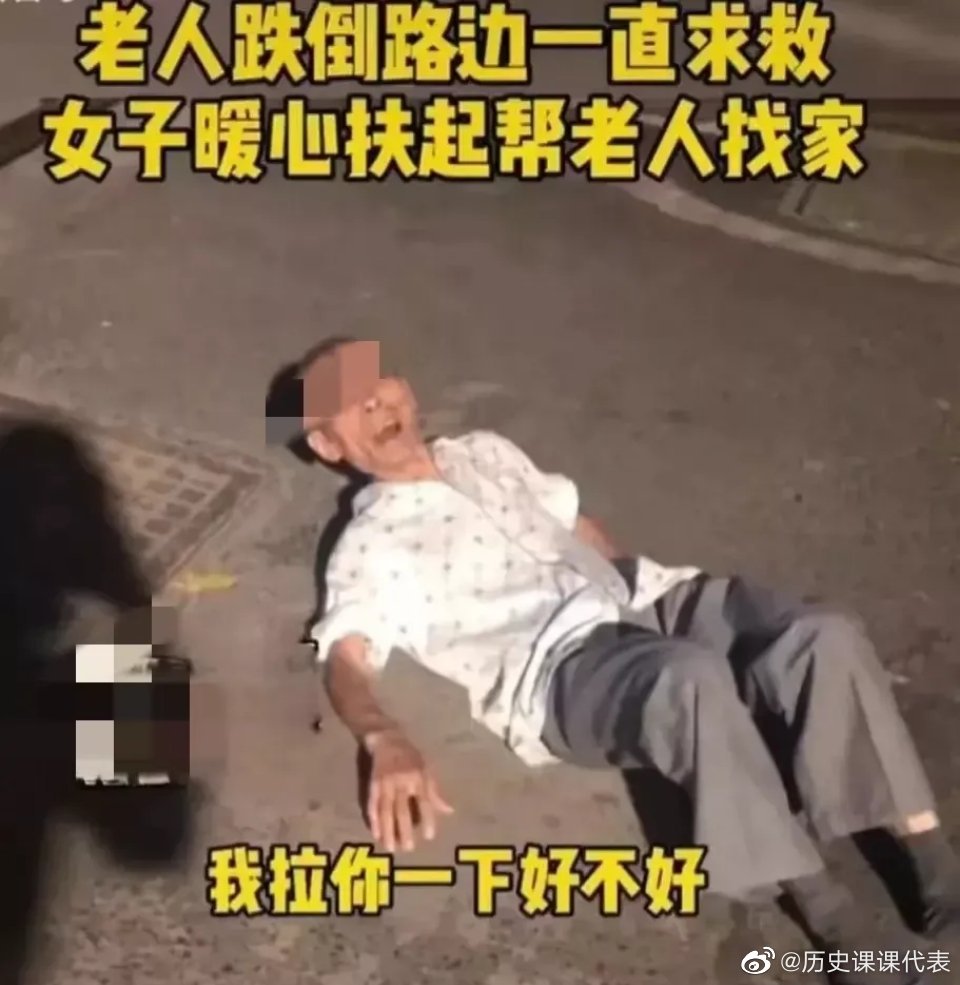} &  \includegraphics[width=3.5cm, height=3.5cm]{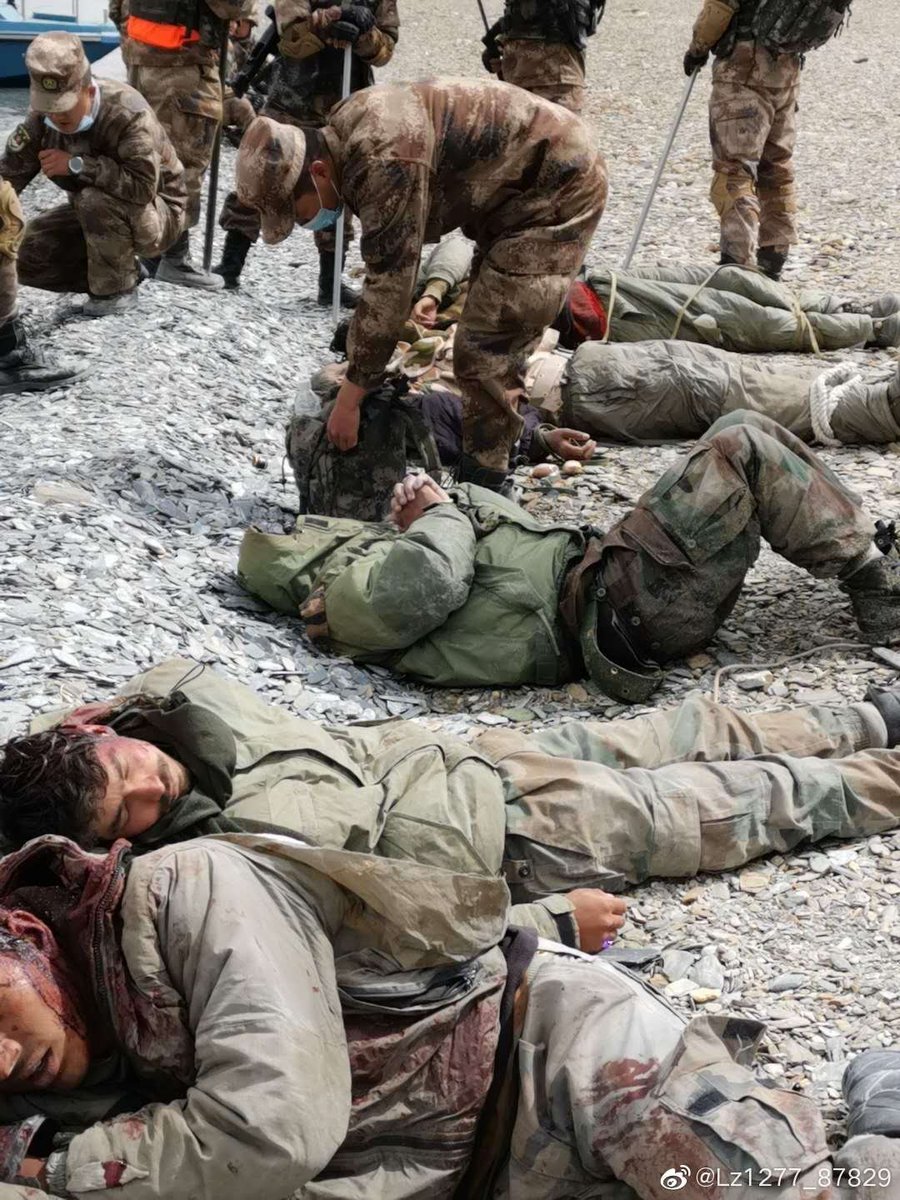} \\
\linespread{0.5}\selectfont
& \begin{CJK}{UTF8}{gbsn}
天津一女子在路上扶起了摔倒的老人帮忙送去救治并垫付3000元，老人清醒后指认是女子撞到了他，老人的子女要求女子再拿10000元。女子报了警...
\end{CJK} (A woman in Tianjin helped an elderly person who had fallen on the road, took him for medical treatment, and paid 3,000 yuan. After the elderly person regained consciousness, he accused the woman of causing the accident. The elderly person's children demanded an additional 10,000 yuan from the woman. The woman reported the incident to the police.) & While India is in denial about border tensions in Ladakh, China linked accounts shared this photo with the caption India China border, in which Indian soldiers are seen lying on the ground with hands and legs tied up with rope.
\\ \midrule
\bf Instruction & Now give you a multimodal post and the corresponding textual evidences... {\color{blue} image ocr}:\begin{CJK}{UTF8}{gbsn}
老人跌倒路边一直求救女子暖心扶起帮老人找家我拉你一下好不好@历史课课代表, 
\end{CJK} (The elderly person fell by the roadside, continuously calling for help. The compassionate woman helped the elderly person up and assisted in finding his way home, saying, "Let me give you a hand, is that okay?" @HistoryClassRep) {\color{blue} image caption}:an old man lying on the ground with his head down... {\color{blue} Evidences}... <and> \begin{CJK}{UTF8}{gbsn} 雷音振长空01-0220:57 {\color{brown} 微博轻享版任何值得称道的正义之光，都应建立在事实之上。可惜的是，这个被这么多大v转发的新闻是假的} \end{CJK} {\color{brown} (Any commendable light of justice should be built upon facts. Unfortunately, the news widely shared by many influencers is fake.)} ... Now,you are required to explain... & Now give you a multimodal post and the corresponding textual evidences... {\color{blue} image ocr}: @Lz1277 87829, {\color{blue} image caption}: soldiers lay on the ground while others stand around them. {\color{blue} Evidences}... <and> {\color{brown} 72 Indian soldiers injured, captured in clash with Chinese Army... <and> Footage claims to show 'Indian troops beating a captured Chinese soldier' amid border tension...} \#\# Explanation: Now, you are required to explain and label the post in three-way classification scheme: ["non-rumor", "rumor", "unverified"].
\\ \midrule
\bf gpt-3.5-turbo 
& The post is labeled as rumor because the textual evidences show conflicting information and the image caption does not directly support the text. {\color{purple} The textual evidence includes multiple instances of the post being shared on social media, but also includes a comment suggesting that the news is fake.} Additionally, the image OCR does not provide any relevant information to confirm the story in the text. Fine-grained labels that are suitable for the post include "Misleading and False" and "Fake News" as the story is being disputed and there is a suggestion that it is not true. Therefore, the post is labeled as <label> rumor </label>. 
& The post is labeled as a rumor because the textual evidence mentions that the photo was circulated among Chinese military websites, and there are conflicting reports about the incident, {\color{purple} with some claiming that Indian soldiers were captured and others claiming that Indian troops beat a captured Chinese soldier. This indicates that there is uncertainty and conflicting information surrounding the photo, leading to its classification as a rumor.} As for the fine-grained labels, "Altered Image" and "Misleading" would be suitable for the post, as the photo is being shared with a misleading caption and there are conflicting reports about the incident depicted in the image. Therefore, the post is labeled as <label> rumor </label>.
\\ \midrule
\bf $\text{\mmidr}_{\text{LLaMA2}}$
& The post is labeled as rumor because the textual and image evidences do not provide clear and credible information about the incident. The textual evidence includes repeated mentions of "Sina Visitor System," which is not relevant to the content of the post. {\color{purple} The image evidence includes a screenshot of a text message in Chinese, which does not provide any relevant information about the incident. Additionally, the image caption describes an old man lying on the ground, which is not directly related to the claim made in the post.} Fine-grained labels that are suitable for the post include "Misleading," "Lacks Context," and "Unverified," as the post lacks clear and credible information, and the image evidence does not support the claim made in the text. Therefore, the post is labeled as <label> rumor </label>.
& The post is labeled as a rumor because the textual evidence suggests that the photo of Indian soldiers being tied up with rope is being circulated among Chinese military websites, {\color{purple} but there is no concrete evidence to support this claim. Additionally, the image OCR and caption do not provide any relevant information to verify the claim made in the text.} Fine-grained labels that are suitable for the post include: Altered Image, Misleading, Misattributed, and False. Therefore, the post is labeled as <label> rumor </label>.
\\ \midrule
\bf  $\text{\mmidr}_{\text{MiniGPT-v2}}$
& The post is labeled as a rumor because the textual evidence includes multiple screenshots of Chinese social media posts and text messages, {\color{purple} but there is no concrete evidence or credible source to support the claim that the woman in the post helped an old man and paid 3000 yuan for his medical expenses. The image caption of an old man lying on the ground with his head down also does not provide any concrete evidence to support the claim.} Fine-grained labels that are suitable for the post include "Misleading", "Lacks Context", and "Unverified". These labels reflect the lack of concrete evidence and the potential for misleading information in the post. Therefore, the post is labeled as <label> rumor </label>.
& The post is labeled as rumor because the textual and image evidences suggest that there is a claim of Indian soldiers being tied up and beaten by Chinese soldiers, which is not verified. {\color{purple} The textual evidence mentions a photo circulating among Chinese military websites showing PLA troops bringing down several Indian soldiers at a close-quarters fight at Pangong Lake border area with India. The image evidence does not directly support the claim, but it does not refute it either.} Fine-grained labels that are suitable for the post include "Misleading", "Misrepresented", and "Unverified". These labels reflect the lack of verification and the potential for misleading information in the post. Therefore, the post is labeled as <label> rumor </label>
\\ \bottomrule
\end{tabular}
}
\label{tab:case_study}  
\end{minipage}
\end{table*}

\subsection{RQ3: Faithful Detection and Interpretable Results}
To evaluate the ability of LLMs in producing high-quality and coherent text explanations for multimodal misinformation, we conducted a case study, as detailed in Table \ref{tab:case_study}. We specifically chose one Chinese example and one English example for in-depth analysis. 
\begin{itemize}[left=0em]
    \item \textbf{Chinese Example (left)}: In this case, the image depicts an elderly man lying on the road. The accompanying text narrates the story of a woman who, driven by goodwill, comes to the aid of the elderly individual. Unfortunately, her good intentions lead to false accusations, compelling her to seek legal intervention by reporting the incident to the police. The most notable evidence includes a comment stating, \textit{"Unfortunately, the news widely shared by many influencers is fake"}.

    \quad It is noteworthy that gpt-3.5-turbo successfully captured the crucial evidence mentioned above, pointing out \textit{"a comment suggesting that the news is fake"}. This demonstrates its exceptional expressive capabilities and comprehensiveness in serving as a teaching model. In contrast, $\text{\mmidr}_{\text{LLaMA2}}$ simply highlights the insufficiency of evidence, acknowledging that the image solely depicts an elderly individual on the ground, without corresponding to the textual description of a female figure. $\text{\mmidr}_{\text{MiniGPT-v2}}$ predominantly raises concerns about the payment of 3000 yuan by the female, concurrently suggesting that the image merely depicts the elderly person on the ground, failing to substantiate the events adequately described in the textual account. This finding indicates that, despite the fact that the distilled student model can generate fluent expressions and provide explanations to some extent, its ability to capture crucial information remains inferior to that of the teacher model.
    
    \item \textbf{English Example (right)}: In this case, the image depicts a group of soldiers lying on the ground. The textual context specifies the location as the India-China border, accompanying an image featuring Indian soldiers lying on the ground, with their hands and legs tied with ropes. The most significant evidence is someone claiming that these are Indian soldiers who have been captured, while others assert that it is the Indian army beating a captured Chinese soldier, presenting conflicting narratives.

    \quad The crucial contradiction mentioned above is successfully captured by gpt-3.5-turbo, further demonstrating its ability to comprehend information. In comparison, $\text{\mmidr}_{\text{LLaMA2}}$ still deems there is insufficient evidence. $\text{\mmidr}_{\text{MiniGPT-v2}}$ asserts a general lack of verification and the potential presence of misleading information. However, it specifically references certain textual evidence related to the India-China conflict, which exhibits limited relevance to the content of the present post. This observation further highlights the remaining gap in explanatory ability of the distilled student model, providing a potential avenue for exploration in future research.
\end{itemize}

\section{Conclusion}
In this paper, we propose \mmidr, a framework designed to teach LLMs to provide fluent and high-quality textual explanations for their assessments. Given multimodal information, we first conduct data augmentation to convert the multimodal retrieval-enhanced misinformation into an appropriate instruction-following format. Subsequently, we feed the processed content into ChatGPT for the extraction of rationales. Finally, we design an efficient knowledge distillation approach, which employs LoRA to train student LLMs (e.g., LLaMA, MiniGPT-v2) on our constructed instruction-following multimodal misinformation dataset. Experimental results demonstrate that our MMIDR exhibits sufficient detection performance and possesses the capacity to generate compelling rationales to interpret its decision-making processes. Nevertheless, the distilled student model still falls short in capturing crucial information compared to the teacher model. We hope our work can facilitate the effective utilization of LLMs and MLLMs in detecting multimodal misinformation.


\bibliographystyle{ACM-Reference-Format}
\bibliography{sample-base}


\begin{thebibliography}{69}


\ifx \showCODEN    \undefined \def \showCODEN     #1{\unskip}     \fi
\ifx \showDOI      \undefined \def \showDOI       #1{#1}\fi
\ifx \showISBNx    \undefined \def \showISBNx     #1{\unskip}     \fi
\ifx \showISBNxiii \undefined \def \showISBNxiii  #1{\unskip}     \fi
\ifx \showISSN     \undefined \def \showISSN      #1{\unskip}     \fi
\ifx \showLCCN     \undefined \def \showLCCN      #1{\unskip}     \fi
\ifx \shownote     \undefined \def \shownote      #1{#1}          \fi
\ifx \showarticletitle \undefined \def \showarticletitle #1{#1}   \fi
\ifx \showURL      \undefined \def \showURL       {\relax}        \fi
\providecommand\bibfield[2]{#2}
\providecommand\bibinfo[2]{#2}
\providecommand\natexlab[1]{#1}
\providecommand\showeprint[2][]{arXiv:#2}

\bibitem[Abdali(2022)]%
        {abdali2022multi}
\bibfield{author}{\bibinfo{person}{Sara Abdali}.}
  \bibinfo{year}{2022}\natexlab{}.
\newblock \showarticletitle{Multi-modal misinformation detection: Approaches,
  challenges and opportunities}.
\newblock \bibinfo{journal}{\emph{arXiv preprint arXiv:2203.13883}}
  (\bibinfo{year}{2022}).
\newblock


\bibitem[Bang et~al\mbox{.}(2023)]%
        {bang2023multitask}
\bibfield{author}{\bibinfo{person}{Yejin Bang}, \bibinfo{person}{Samuel
  Cahyawijaya}, \bibinfo{person}{Nayeon Lee}, \bibinfo{person}{Wenliang Dai},
  \bibinfo{person}{Dan Su}, \bibinfo{person}{Bryan Wilie},
  \bibinfo{person}{Holy Lovenia}, \bibinfo{person}{Ziwei Ji},
  \bibinfo{person}{Tiezheng Yu}, \bibinfo{person}{Willy Chung},
  {et~al\mbox{.}}} \bibinfo{year}{2023}\natexlab{}.
\newblock \showarticletitle{A multitask, multilingual, multimodal evaluation of
  chatgpt on reasoning, hallucination, and interactivity}.
\newblock \bibinfo{journal}{\emph{arXiv preprint arXiv:2302.04023}}
  (\bibinfo{year}{2023}).
\newblock


\bibitem[Bhattarai et~al\mbox{.}(2021)]%
        {bhattarai2021explainable}
\bibfield{author}{\bibinfo{person}{Bimal Bhattarai},
  \bibinfo{person}{Ole-Christoffer Granmo}, {and} \bibinfo{person}{Lei Jiao}.}
  \bibinfo{year}{2021}\natexlab{}.
\newblock \showarticletitle{Explainable tsetlin machine framework for fake news
  detection with credibility score assessment}.
\newblock \bibinfo{journal}{\emph{arXiv preprint arXiv:2105.09114}}
  (\bibinfo{year}{2021}).
\newblock


\bibitem[Brown et~al\mbox{.}(2020)]%
        {brown2020language}
\bibfield{author}{\bibinfo{person}{Tom Brown}, \bibinfo{person}{Benjamin Mann},
  \bibinfo{person}{Nick Ryder}, \bibinfo{person}{Melanie Subbiah},
  \bibinfo{person}{Jared~D Kaplan}, \bibinfo{person}{Prafulla Dhariwal},
  \bibinfo{person}{Arvind Neelakantan}, \bibinfo{person}{Pranav Shyam},
  \bibinfo{person}{Girish Sastry}, \bibinfo{person}{Amanda Askell},
  {et~al\mbox{.}}} \bibinfo{year}{2020}\natexlab{}.
\newblock \showarticletitle{Language models are few-shot learners}.
\newblock \bibinfo{journal}{\emph{Advances in neural information processing
  systems}}  \bibinfo{volume}{33} (\bibinfo{year}{2020}),
  \bibinfo{pages}{1877--1901}.
\newblock


\bibitem[Buchholz(2023)]%
        {buchholz2023assessing}
\bibfield{author}{\bibinfo{person}{Mars~Gokturk Buchholz}.}
  \bibinfo{year}{2023}\natexlab{}.
\newblock \showarticletitle{Assessing the Effectiveness of GPT-3 in Detecting
  False Political Statements: A Case Study on the LIAR Dataset}.
\newblock \bibinfo{journal}{\emph{arXiv preprint arXiv:2306.08190}}
  (\bibinfo{year}{2023}).
\newblock


\bibitem[Cao et~al\mbox{.}(2020)]%
        {cao2020exploring}
\bibfield{author}{\bibinfo{person}{Juan Cao}, \bibinfo{person}{Peng Qi},
  \bibinfo{person}{Qiang Sheng}, \bibinfo{person}{Tianyun Yang},
  \bibinfo{person}{Junbo Guo}, {and} \bibinfo{person}{Jintao Li}.}
  \bibinfo{year}{2020}\natexlab{}.
\newblock \showarticletitle{Exploring the role of visual content in fake news
  detection}.
\newblock \bibinfo{journal}{\emph{Disinformation, Misinformation, and Fake News
  in Social Media: Emerging Research Challenges and Opportunities}}
  (\bibinfo{year}{2020}), \bibinfo{pages}{141--161}.
\newblock


\bibitem[Caramancion(2023)]%
        {caramancion2023harnessing}
\bibfield{author}{\bibinfo{person}{Kevin~Matthe Caramancion}.}
  \bibinfo{year}{2023}\natexlab{}.
\newblock \showarticletitle{Harnessing the Power of ChatGPT to Decimate
  Mis/Disinformation: Using ChatGPT for Fake News Detection}. In
  \bibinfo{booktitle}{\emph{2023 IEEE World AI IoT Congress (AIIoT)}}. IEEE,
  \bibinfo{pages}{0042--0046}.
\newblock


\bibitem[Chen and Shu(2023)]%
        {chen2023combating}
\bibfield{author}{\bibinfo{person}{Canyu Chen} {and} \bibinfo{person}{Kai
  Shu}.} \bibinfo{year}{2023}\natexlab{}.
\newblock \showarticletitle{Combating misinformation in the age of llms:
  Opportunities and challenges}.
\newblock \bibinfo{journal}{\emph{arXiv preprint arXiv:2311.05656}}
  (\bibinfo{year}{2023}).
\newblock


\bibitem[Chen et~al\mbox{.}(2023)]%
        {chen2023minigpt}
\bibfield{author}{\bibinfo{person}{Jun Chen}, \bibinfo{person}{Deyao Zhu},
  \bibinfo{person}{Xiaoqian Shen}, \bibinfo{person}{Xiang Li},
  \bibinfo{person}{Zechun Liu}, \bibinfo{person}{Pengchuan Zhang},
  \bibinfo{person}{Raghuraman Krishnamoorthi}, \bibinfo{person}{Vikas Chandra},
  \bibinfo{person}{Yunyang Xiong}, {and} \bibinfo{person}{Mohamed Elhoseiny}.}
  \bibinfo{year}{2023}\natexlab{}.
\newblock \showarticletitle{Minigpt-v2: large language model as a unified
  interface for vision-language multi-task learning}.
\newblock \bibinfo{journal}{\emph{arXiv preprint arXiv:2310.09478}}
  (\bibinfo{year}{2023}).
\newblock


\bibitem[Chen et~al\mbox{.}(2022)]%
        {chen2022cross}
\bibfield{author}{\bibinfo{person}{Yixuan Chen}, \bibinfo{person}{Dongsheng
  Li}, \bibinfo{person}{Peng Zhang}, \bibinfo{person}{Jie Sui},
  \bibinfo{person}{Qin Lv}, \bibinfo{person}{Lu Tun}, {and} \bibinfo{person}{Li
  Shang}.} \bibinfo{year}{2022}\natexlab{}.
\newblock \showarticletitle{Cross-modal ambiguity learning for multimodal fake
  news detection}. In \bibinfo{booktitle}{\emph{Proceedings of the ACM Web
  Conference 2022}}. \bibinfo{pages}{2897--2905}.
\newblock


\bibitem[Chern et~al\mbox{.}(2023)]%
        {chern2023factool}
\bibfield{author}{\bibinfo{person}{I Chern}, \bibinfo{person}{Steffi Chern},
  \bibinfo{person}{Shiqi Chen}, \bibinfo{person}{Weizhe Yuan},
  \bibinfo{person}{Kehua Feng}, \bibinfo{person}{Chunting Zhou},
  \bibinfo{person}{Junxian He}, \bibinfo{person}{Graham Neubig},
  \bibinfo{person}{Pengfei Liu}, {et~al\mbox{.}}}
  \bibinfo{year}{2023}\natexlab{}.
\newblock \showarticletitle{FacTool: Factuality Detection in Generative AI--A
  Tool Augmented Framework for Multi-Task and Multi-Domain Scenarios}.
\newblock \bibinfo{journal}{\emph{arXiv preprint arXiv:2307.13528}}
  (\bibinfo{year}{2023}).
\newblock


\bibitem[Cheung and Lam(2023)]%
        {cheung2023factllama}
\bibfield{author}{\bibinfo{person}{Tsun-Hin Cheung} {and}
  \bibinfo{person}{Kin-Man Lam}.} \bibinfo{year}{2023}\natexlab{}.
\newblock \showarticletitle{FactLLaMA: Optimizing Instruction-Following
  Language Models with External Knowledge for Automated Fact-Checking}. In
  \bibinfo{booktitle}{\emph{2023 Asia Pacific Signal and Information Processing
  Association Annual Summit and Conference (APSIPA ASC)}}. IEEE,
  \bibinfo{pages}{846--853}.
\newblock


\bibitem[Conroy et~al\mbox{.}(2015)]%
        {conroy2015automatic}
\bibfield{author}{\bibinfo{person}{Nadia~K Conroy}, \bibinfo{person}{Victoria~L
  Rubin}, {and} \bibinfo{person}{Yimin Chen}.} \bibinfo{year}{2015}\natexlab{}.
\newblock \showarticletitle{Automatic deception detection: Methods for finding
  fake news}.
\newblock \bibinfo{journal}{\emph{Proceedings of the association for
  information science and technology}} \bibinfo{volume}{52},
  \bibinfo{number}{1} (\bibinfo{year}{2015}), \bibinfo{pages}{1--4}.
\newblock


\bibitem[Devlin et~al\mbox{.}(2018)]%
        {devlin2018bert}
\bibfield{author}{\bibinfo{person}{Jacob Devlin}, \bibinfo{person}{Ming-Wei
  Chang}, \bibinfo{person}{Kenton Lee}, {and} \bibinfo{person}{Kristina
  Toutanova}.} \bibinfo{year}{2018}\natexlab{}.
\newblock \showarticletitle{Bert: Pre-training of deep bidirectional
  transformers for language understanding}.
\newblock \bibinfo{journal}{\emph{arXiv preprint arXiv:1810.04805}}
  (\bibinfo{year}{2018}).
\newblock


\bibitem[Fang et~al\mbox{.}(2023)]%
        {Fang_2023_CVPR}
\bibfield{author}{\bibinfo{person}{Yuxin Fang}, \bibinfo{person}{Wen Wang},
  \bibinfo{person}{Binhui Xie}, \bibinfo{person}{Quan Sun},
  \bibinfo{person}{Ledell Wu}, \bibinfo{person}{Xinggang Wang},
  \bibinfo{person}{Tiejun Huang}, \bibinfo{person}{Xinlong Wang}, {and}
  \bibinfo{person}{Yue Cao}.} \bibinfo{year}{2023}\natexlab{}.
\newblock \showarticletitle{EVA: Exploring the Limits of Masked Visual
  Representation Learning at Scale}. In \bibinfo{booktitle}{\emph{Proceedings
  of the IEEE/CVF Conference on Computer Vision and Pattern Recognition
  (CVPR)}}. \bibinfo{pages}{19358--19369}.
\newblock


\bibitem[Gou et~al\mbox{.}(2021)]%
        {gou2021knowledge}
\bibfield{author}{\bibinfo{person}{Jianping Gou}, \bibinfo{person}{Baosheng
  Yu}, \bibinfo{person}{Stephen~J Maybank}, {and} \bibinfo{person}{Dacheng
  Tao}.} \bibinfo{year}{2021}\natexlab{}.
\newblock \showarticletitle{Knowledge distillation: A survey}.
\newblock \bibinfo{journal}{\emph{International Journal of Computer Vision}}
  \bibinfo{volume}{129}, \bibinfo{number}{6} (\bibinfo{year}{2021}),
  \bibinfo{pages}{1789--1819}.
\newblock


\bibitem[Guo et~al\mbox{.}(2018)]%
        {guo2018rumor}
\bibfield{author}{\bibinfo{person}{Han Guo}, \bibinfo{person}{Juan Cao},
  \bibinfo{person}{Yazi Zhang}, \bibinfo{person}{Junbo Guo}, {and}
  \bibinfo{person}{Jintao Li}.} \bibinfo{year}{2018}\natexlab{}.
\newblock \showarticletitle{Rumor detection with hierarchical social attention
  network}. In \bibinfo{booktitle}{\emph{Proceedings of the 27th ACM
  international conference on information and knowledge management}}.
  \bibinfo{pages}{943--951}.
\newblock


\bibitem[Gupta and Agrawal(2022)]%
        {gupta2022compression}
\bibfield{author}{\bibinfo{person}{Manish Gupta} {and} \bibinfo{person}{Puneet
  Agrawal}.} \bibinfo{year}{2022}\natexlab{}.
\newblock \showarticletitle{Compression of deep learning models for text: A
  survey}.
\newblock \bibinfo{journal}{\emph{ACM Transactions on Knowledge Discovery from
  Data (TKDD)}} \bibinfo{volume}{16}, \bibinfo{number}{4}
  (\bibinfo{year}{2022}), \bibinfo{pages}{1--55}.
\newblock


\bibitem[He et~al\mbox{.}(2016)]%
        {he2016deep}
\bibfield{author}{\bibinfo{person}{Kaiming He}, \bibinfo{person}{Xiangyu
  Zhang}, \bibinfo{person}{Shaoqing Ren}, {and} \bibinfo{person}{Jian Sun}.}
  \bibinfo{year}{2016}\natexlab{}.
\newblock \showarticletitle{Deep residual learning for image recognition}. In
  \bibinfo{booktitle}{\emph{Proceedings of the IEEE conference on computer
  vision and pattern recognition}}. \bibinfo{pages}{770--778}.
\newblock


\bibitem[Hsieh et~al\mbox{.}(2023)]%
        {hsieh2023distilling}
\bibfield{author}{\bibinfo{person}{Cheng{-}Yu Hsieh},
  \bibinfo{person}{Chun{-}Liang Li}, \bibinfo{person}{Chih{-}Kuan Yeh},
  \bibinfo{person}{Hootan Nakhost}, \bibinfo{person}{Yasuhisa Fujii},
  \bibinfo{person}{Alex Ratner}, \bibinfo{person}{Ranjay Krishna},
  \bibinfo{person}{Chen{-}Yu Lee}, {and} \bibinfo{person}{Tomas Pfister}.}
  \bibinfo{year}{2023}\natexlab{}.
\newblock \showarticletitle{Distilling Step-by-Step! Outperforming Larger
  Language Models with Less Training Data and Smaller Model Sizes}. In
  \bibinfo{booktitle}{\emph{{ACL} (Findings)}}. \bibinfo{publisher}{Association
  for Computational Linguistics}, \bibinfo{pages}{8003--8017}.
\newblock


\bibitem[Hu et~al\mbox{.}(2023b)]%
        {hu2023bad}
\bibfield{author}{\bibinfo{person}{Beizhe Hu}, \bibinfo{person}{Qiang Sheng},
  \bibinfo{person}{Juan Cao}, \bibinfo{person}{Yuhui Shi},
  \bibinfo{person}{Yang Li}, \bibinfo{person}{Danding Wang}, {and}
  \bibinfo{person}{Peng Qi}.} \bibinfo{year}{2023}\natexlab{b}.
\newblock \showarticletitle{Bad actor, good advisor: Exploring the role of
  large language models in fake news detection}.
\newblock \bibinfo{journal}{\emph{arXiv preprint arXiv:2309.12247}}
  (\bibinfo{year}{2023}).
\newblock


\bibitem[Hu et~al\mbox{.}(2021)]%
        {hu2021lora}
\bibfield{author}{\bibinfo{person}{Edward~J Hu}, \bibinfo{person}{Yelong Shen},
  \bibinfo{person}{Phillip Wallis}, \bibinfo{person}{Zeyuan Allen-Zhu},
  \bibinfo{person}{Yuanzhi Li}, \bibinfo{person}{Shean Wang},
  \bibinfo{person}{Lu Wang}, {and} \bibinfo{person}{Weizhu Chen}.}
  \bibinfo{year}{2021}\natexlab{}.
\newblock \showarticletitle{Lora: Low-rank adaptation of large language
  models}.
\newblock \bibinfo{journal}{\emph{arXiv preprint arXiv:2106.09685}}
  (\bibinfo{year}{2021}).
\newblock


\bibitem[Hu et~al\mbox{.}(2022)]%
        {hu2022deep}
\bibfield{author}{\bibinfo{person}{LinMei Hu}, \bibinfo{person}{SiQi Wei},
  \bibinfo{person}{Ziwang Zhao}, {and} \bibinfo{person}{Bin Wu}.}
  \bibinfo{year}{2022}\natexlab{}.
\newblock \showarticletitle{Deep learning for fake news detection: A
  comprehensive survey}.
\newblock \bibinfo{journal}{\emph{AI Open}} (\bibinfo{year}{2022}).
\newblock


\bibitem[Hu et~al\mbox{.}(2023a)]%
        {hu2023mr2}
\bibfield{author}{\bibinfo{person}{Xuming Hu}, \bibinfo{person}{Zhijiang Guo},
  \bibinfo{person}{Junzhe Chen}, \bibinfo{person}{Lijie Wen}, {and}
  \bibinfo{person}{Philip~S Yu}.} \bibinfo{year}{2023}\natexlab{a}.
\newblock \showarticletitle{Mr2: A benchmark for multimodal retrieval-augmented
  rumor detection in social media}. In \bibinfo{booktitle}{\emph{Proceedings of
  the 46th international ACM SIGIR conference on research and development in
  information retrieval}}. \bibinfo{pages}{2901--2912}.
\newblock


\bibitem[Jiang et~al\mbox{.}(2023)]%
        {jiang2023raucg}
\bibfield{author}{\bibinfo{person}{Shuyu Jiang}, \bibinfo{person}{Wenyi Tang},
  \bibinfo{person}{Xingshu Chen}, \bibinfo{person}{Rui Tanga},
  \bibinfo{person}{Haizhou Wang}, {and} \bibinfo{person}{Wenxian Wang}.}
  \bibinfo{year}{2023}\natexlab{}.
\newblock \showarticletitle{Raucg: Retrieval-augmented unsupervised counter
  narrative generation for hate speech}.
\newblock \bibinfo{journal}{\emph{arXiv preprint arXiv:2310.05650}}
  (\bibinfo{year}{2023}).
\newblock


\bibitem[Jung et~al\mbox{.}(2023)]%
        {jung2023impossible}
\bibfield{author}{\bibinfo{person}{Jaehun Jung}, \bibinfo{person}{Peter West},
  \bibinfo{person}{Liwei Jiang}, \bibinfo{person}{Faeze Brahman},
  \bibinfo{person}{Ximing Lu}, \bibinfo{person}{Jillian Fisher},
  \bibinfo{person}{Taylor Sorensen}, {and} \bibinfo{person}{Yejin Choi}.}
  \bibinfo{year}{2023}\natexlab{}.
\newblock \bibinfo{title}{Impossible Distillation: from Low-Quality Model to
  High-Quality Dataset \& Model for Summarization and Paraphrasing}.
\newblock
\newblock
\showeprint[arxiv]{2305.16635}~[cs.CL]


\bibitem[Khattar et~al\mbox{.}(2019)]%
        {khattar2019mvae}
\bibfield{author}{\bibinfo{person}{Dhruv Khattar},
  \bibinfo{person}{Jaipal~Singh Goud}, \bibinfo{person}{Manish Gupta}, {and}
  \bibinfo{person}{Vasudeva Varma}.} \bibinfo{year}{2019}\natexlab{}.
\newblock \showarticletitle{Mvae: Multimodal variational autoencoder for fake
  news detection}. In \bibinfo{booktitle}{\emph{The world wide web
  conference}}. \bibinfo{pages}{2915--2921}.
\newblock


\bibitem[Li et~al\mbox{.}(2023a)]%
        {li2023blip}
\bibfield{author}{\bibinfo{person}{Junnan Li}, \bibinfo{person}{Dongxu Li},
  \bibinfo{person}{Silvio Savarese}, {and} \bibinfo{person}{Steven Hoi}.}
  \bibinfo{year}{2023}\natexlab{a}.
\newblock \showarticletitle{Blip-2: Bootstrapping language-image pre-training
  with frozen image encoders and large language models}.
\newblock \bibinfo{journal}{\emph{arXiv preprint arXiv:2301.12597}}
  (\bibinfo{year}{2023}).
\newblock


\bibitem[Li et~al\mbox{.}(2021)]%
        {DBLP:conf/acl/LiNK21}
\bibfield{author}{\bibinfo{person}{Jiawen Li}, \bibinfo{person}{Shiwen Ni},
  {and} \bibinfo{person}{Hung{-}Yu Kao}.} \bibinfo{year}{2021}\natexlab{}.
\newblock \showarticletitle{Meet The Truth: Leverage Objective Facts and
  Subjective Views for Interpretable Rumor Detection}. In
  \bibinfo{booktitle}{\emph{Findings of the Association for Computational
  Linguistics: {ACL/IJCNLP} 2021, Online Event, August 1-6, 2021}}
  \emph{(\bibinfo{series}{Findings of {ACL}},
  Vol.~\bibinfo{volume}{{ACL/IJCNLP} 2021})},
  \bibfield{editor}{\bibinfo{person}{Chengqing Zong}, \bibinfo{person}{Fei
  Xia}, \bibinfo{person}{Wenjie Li}, {and} \bibinfo{person}{Roberto Navigli}}
  (Eds.). \bibinfo{publisher}{Association for Computational Linguistics},
  \bibinfo{pages}{705--715}.
\newblock
\urldef\tempurl%
\url{https://doi.org/10.18653/V1/2021.FINDINGS-ACL.63}
\showDOI{\tempurl}


\bibitem[Li et~al\mbox{.}(2023b)]%
        {li2023self}
\bibfield{author}{\bibinfo{person}{Miaoran Li}, \bibinfo{person}{Baolin Peng},
  {and} \bibinfo{person}{Zhu Zhang}.} \bibinfo{year}{2023}\natexlab{b}.
\newblock \showarticletitle{Self-Checker: Plug-and-Play Modules for
  Fact-Checking with Large Language Models}.
\newblock \bibinfo{journal}{\emph{arXiv preprint arXiv:2305.14623}}
  (\bibinfo{year}{2023}).
\newblock


\bibitem[Liu et~al\mbox{.}(2024)]%
        {liu2024visual}
\bibfield{author}{\bibinfo{person}{Haotian Liu}, \bibinfo{person}{Chunyuan Li},
  \bibinfo{person}{Qingyang Wu}, {and} \bibinfo{person}{Yong~Jae Lee}.}
  \bibinfo{year}{2024}\natexlab{}.
\newblock \showarticletitle{Visual instruction tuning}.
\newblock \bibinfo{journal}{\emph{Advances in neural information processing
  systems}}  \bibinfo{volume}{36} (\bibinfo{year}{2024}).
\newblock


\bibitem[Liu et~al\mbox{.}(2023)]%
        {liu2023minds}
\bibfield{author}{\bibinfo{person}{Weize Liu}, \bibinfo{person}{Guocong Li},
  \bibinfo{person}{Kai Zhang}, \bibinfo{person}{Bang Du},
  \bibinfo{person}{Qiyuan Chen}, \bibinfo{person}{Xuming Hu},
  \bibinfo{person}{Hongxia Xu}, \bibinfo{person}{Jintai Chen}, {and}
  \bibinfo{person}{Jian Wu}.} \bibinfo{year}{2023}\natexlab{}.
\newblock \bibinfo{title}{Mind's Mirror: Distilling Self-Evaluation Capability
  and Comprehensive Thinking from Large Language Models}.
\newblock
\newblock
\showeprint[arxiv]{2311.09214}~[cs.CL]


\bibitem[Loshchilov and Hutter(2018)]%
        {loshchilov2018decoupled}
\bibfield{author}{\bibinfo{person}{Ilya Loshchilov} {and}
  \bibinfo{person}{Frank Hutter}.} \bibinfo{year}{2018}\natexlab{}.
\newblock \showarticletitle{Decoupled Weight Decay Regularization}. In
  \bibinfo{booktitle}{\emph{International Conference on Learning
  Representations}}.
\newblock


\bibitem[Lu and Li(2020)]%
        {lu2020gcan}
\bibfield{author}{\bibinfo{person}{Yi-Ju Lu} {and} \bibinfo{person}{Cheng-Te
  Li}.} \bibinfo{year}{2020}\natexlab{}.
\newblock \showarticletitle{GCAN: Graph-aware co-attention networks for
  explainable fake news detection on social media}.
\newblock \bibinfo{journal}{\emph{arXiv preprint arXiv:2004.11648}}
  (\bibinfo{year}{2020}).
\newblock


\bibitem[Luo et~al\mbox{.}(2023a)]%
        {luo2023wizardmath}
\bibfield{author}{\bibinfo{person}{Haipeng Luo}, \bibinfo{person}{Qingfeng
  Sun}, \bibinfo{person}{Can Xu}, \bibinfo{person}{Pu Zhao},
  \bibinfo{person}{Jianguang Lou}, \bibinfo{person}{Chongyang Tao},
  \bibinfo{person}{Xiubo Geng}, \bibinfo{person}{Qingwei Lin},
  \bibinfo{person}{Shifeng Chen}, {and} \bibinfo{person}{Dongmei Zhang}.}
  \bibinfo{year}{2023}\natexlab{a}.
\newblock \showarticletitle{Wizardmath: Empowering mathematical reasoning for
  large language models via reinforced evol-instruct}.
\newblock \bibinfo{journal}{\emph{arXiv preprint arXiv:2308.09583}}
  (\bibinfo{year}{2023}).
\newblock


\bibitem[Luo et~al\mbox{.}(2023b)]%
        {luo2023wizardcoder}
\bibfield{author}{\bibinfo{person}{Ziyang Luo}, \bibinfo{person}{Can Xu},
  \bibinfo{person}{Pu Zhao}, \bibinfo{person}{Qingfeng Sun},
  \bibinfo{person}{Xiubo Geng}, \bibinfo{person}{Wenxiang Hu},
  \bibinfo{person}{Chongyang Tao}, \bibinfo{person}{Jing Ma},
  \bibinfo{person}{Qingwei Lin}, {and} \bibinfo{person}{Daxin Jiang}.}
  \bibinfo{year}{2023}\natexlab{b}.
\newblock \showarticletitle{WizardCoder: Empowering Code Large Language Models
  with Evol-Instruct}.
\newblock \bibinfo{journal}{\emph{arXiv preprint arXiv:2306.08568}}
  (\bibinfo{year}{2023}).
\newblock


\bibitem[Nielsen and McConville(2022)]%
        {10.1145/3477495.3531744}
\bibfield{author}{\bibinfo{person}{Dan~S. Nielsen} {and} \bibinfo{person}{Ryan
  McConville}.} \bibinfo{year}{2022}\natexlab{}.
\newblock \showarticletitle{MuMiN: A Large-Scale Multilingual Multimodal
  Fact-Checked Misinformation Social Network Dataset}. In
  \bibinfo{booktitle}{\emph{Proceedings of the 45th International ACM SIGIR
  Conference on Research and Development in Information Retrieval}}
  (<conf-loc>, <city>Madrid</city>, <country>Spain</country>, </conf-loc>)
  \emph{(\bibinfo{series}{SIGIR '22})}. \bibinfo{publisher}{Association for
  Computing Machinery}, \bibinfo{address}{New York, NY, USA},
  \bibinfo{pages}{3141–3153}.
\newblock
\showISBNx{9781450387323}
\urldef\tempurl%
\url{https://doi.org/10.1145/3477495.3531744}
\showDOI{\tempurl}


\bibitem[OpenAI et~al\mbox{.}(2023)]%
        {openai2023gpt4}
\bibfield{author}{\bibinfo{person}{OpenAI}, \bibinfo{person}{:},
  \bibinfo{person}{Josh Achiam}, \bibinfo{person}{Steven Adler},
  \bibinfo{person}{Sandhini Agarwal}, \bibinfo{person}{Lama Ahmad},
  \bibinfo{person}{Ilge Akkaya}, \bibinfo{person}{Florencia~Leoni Aleman},
  \bibinfo{person}{Diogo Almeida}, \bibinfo{person}{Janko Altenschmidt},
  \bibinfo{person}{Sam Altman}, \bibinfo{person}{Shyamal Anadkat},
  \bibinfo{person}{Red Avila}, \bibinfo{person}{Igor Babuschkin},
  \bibinfo{person}{Suchir Balaji}, \bibinfo{person}{Valerie Balcom},
  \bibinfo{person}{Paul Baltescu}, \bibinfo{person}{Haiming Bao},
  \bibinfo{person}{Mo Bavarian}, \bibinfo{person}{Jeff Belgum},
  \bibinfo{person}{Irwan Bello}, \bibinfo{person}{Jake Berdine},
  \bibinfo{person}{Gabriel Bernadett-Shapiro}, \bibinfo{person}{Christopher
  Berner}, \bibinfo{person}{Lenny Bogdonoff}, \bibinfo{person}{Oleg Boiko},
  \bibinfo{person}{Madelaine Boyd}, \bibinfo{person}{Anna-Luisa Brakman},
  \bibinfo{person}{Greg Brockman}, \bibinfo{person}{Tim Brooks},
  \bibinfo{person}{Miles Brundage}, \bibinfo{person}{Kevin Button},
  \bibinfo{person}{Trevor Cai}, \bibinfo{person}{Rosie Campbell},
  \bibinfo{person}{Andrew Cann}, \bibinfo{person}{Brittany Carey},
  \bibinfo{person}{Chelsea Carlson}, \bibinfo{person}{Rory Carmichael},
  \bibinfo{person}{Brooke Chan}, \bibinfo{person}{Che Chang},
  \bibinfo{person}{Fotis Chantzis}, \bibinfo{person}{Derek Chen},
  \bibinfo{person}{Sully Chen}, \bibinfo{person}{Ruby Chen},
  \bibinfo{person}{Jason Chen}, \bibinfo{person}{Mark Chen},
  \bibinfo{person}{Ben Chess}, \bibinfo{person}{Chester Cho},
  \bibinfo{person}{Casey Chu}, \bibinfo{person}{Hyung~Won Chung},
  \bibinfo{person}{Dave Cummings}, \bibinfo{person}{Jeremiah Currier},
  \bibinfo{person}{Yunxing Dai}, \bibinfo{person}{Cory Decareaux},
  \bibinfo{person}{Thomas Degry}, \bibinfo{person}{Noah Deutsch},
  \bibinfo{person}{Damien Deville}, \bibinfo{person}{Arka Dhar},
  \bibinfo{person}{David Dohan}, \bibinfo{person}{Steve Dowling},
  \bibinfo{person}{Sheila Dunning}, \bibinfo{person}{Adrien Ecoffet},
  \bibinfo{person}{Atty Eleti}, \bibinfo{person}{Tyna Eloundou},
  \bibinfo{person}{David Farhi}, \bibinfo{person}{Liam Fedus},
  \bibinfo{person}{Niko Felix}, \bibinfo{person}{Simón~Posada Fishman},
  \bibinfo{person}{Juston Forte}, \bibinfo{person}{Isabella Fulford},
  \bibinfo{person}{Leo Gao}, \bibinfo{person}{Elie Georges},
  \bibinfo{person}{Christian Gibson}, \bibinfo{person}{Vik Goel},
  \bibinfo{person}{Tarun Gogineni}, \bibinfo{person}{Gabriel Goh},
  \bibinfo{person}{Rapha Gontijo-Lopes}, \bibinfo{person}{Jonathan Gordon},
  \bibinfo{person}{Morgan Grafstein}, \bibinfo{person}{Scott Gray},
  \bibinfo{person}{Ryan Greene}, \bibinfo{person}{Joshua Gross},
  \bibinfo{person}{Shixiang~Shane Gu}, \bibinfo{person}{Yufei Guo},
  \bibinfo{person}{Chris Hallacy}, \bibinfo{person}{Jesse Han},
  \bibinfo{person}{Jeff Harris}, \bibinfo{person}{Yuchen He},
  \bibinfo{person}{Mike Heaton}, \bibinfo{person}{Johannes Heidecke},
  \bibinfo{person}{Chris Hesse}, \bibinfo{person}{Alan Hickey},
  \bibinfo{person}{Wade Hickey}, \bibinfo{person}{Peter Hoeschele},
  \bibinfo{person}{Brandon Houghton}, \bibinfo{person}{Kenny Hsu},
  \bibinfo{person}{Shengli Hu}, \bibinfo{person}{Xin Hu},
  \bibinfo{person}{Joost Huizinga}, \bibinfo{person}{Shantanu Jain},
  \bibinfo{person}{Shawn Jain}, \bibinfo{person}{Joanne Jang},
  \bibinfo{person}{Angela Jiang}, \bibinfo{person}{Roger Jiang},
  \bibinfo{person}{Haozhun Jin}, \bibinfo{person}{Denny Jin},
  \bibinfo{person}{Shino Jomoto}, \bibinfo{person}{Billie Jonn},
  \bibinfo{person}{Heewoo Jun}, \bibinfo{person}{Tomer Kaftan},
  \bibinfo{person}{Łukasz Kaiser}, \bibinfo{person}{Ali Kamali},
  \bibinfo{person}{Ingmar Kanitscheider}, \bibinfo{person}{Nitish~Shirish
  Keskar}, \bibinfo{person}{Tabarak Khan}, \bibinfo{person}{Logan Kilpatrick},
  \bibinfo{person}{Jong~Wook Kim}, \bibinfo{person}{Christina Kim},
  \bibinfo{person}{Yongjik Kim}, \bibinfo{person}{Hendrik Kirchner},
  \bibinfo{person}{Jamie Kiros}, \bibinfo{person}{Matt Knight},
  \bibinfo{person}{Daniel Kokotajlo}, \bibinfo{person}{Łukasz Kondraciuk},
  \bibinfo{person}{Andrew Kondrich}, \bibinfo{person}{Aris Konstantinidis},
  \bibinfo{person}{Kyle Kosic}, \bibinfo{person}{Gretchen Krueger},
  \bibinfo{person}{Vishal Kuo}, \bibinfo{person}{Michael Lampe},
  \bibinfo{person}{Ikai Lan}, \bibinfo{person}{Teddy Lee}, \bibinfo{person}{Jan
  Leike}, \bibinfo{person}{Jade Leung}, \bibinfo{person}{Daniel Levy},
  \bibinfo{person}{Chak~Ming Li}, \bibinfo{person}{Rachel Lim},
  \bibinfo{person}{Molly Lin}, \bibinfo{person}{Stephanie Lin},
  \bibinfo{person}{Mateusz Litwin}, \bibinfo{person}{Theresa Lopez},
  \bibinfo{person}{Ryan Lowe}, \bibinfo{person}{Patricia Lue},
  \bibinfo{person}{Anna Makanju}, \bibinfo{person}{Kim Malfacini},
  \bibinfo{person}{Sam Manning}, \bibinfo{person}{Todor Markov},
  \bibinfo{person}{Yaniv Markovski}, \bibinfo{person}{Bianca Martin},
  \bibinfo{person}{Katie Mayer}, \bibinfo{person}{Andrew Mayne},
  \bibinfo{person}{Bob McGrew}, \bibinfo{person}{Scott~Mayer McKinney},
  \bibinfo{person}{Christine McLeavey}, \bibinfo{person}{Paul McMillan},
  \bibinfo{person}{Jake McNeil}, \bibinfo{person}{David Medina},
  \bibinfo{person}{Aalok Mehta}, \bibinfo{person}{Jacob Menick},
  \bibinfo{person}{Luke Metz}, \bibinfo{person}{Andrey Mishchenko},
  \bibinfo{person}{Pamela Mishkin}, \bibinfo{person}{Vinnie Monaco},
  \bibinfo{person}{Evan Morikawa}, \bibinfo{person}{Daniel Mossing},
  \bibinfo{person}{Tong Mu}, \bibinfo{person}{Mira Murati},
  \bibinfo{person}{Oleg Murk}, \bibinfo{person}{David Mély},
  \bibinfo{person}{Ashvin Nair}, \bibinfo{person}{Reiichiro Nakano},
  \bibinfo{person}{Rajeev Nayak}, \bibinfo{person}{Arvind Neelakantan},
  \bibinfo{person}{Richard Ngo}, \bibinfo{person}{Hyeonwoo Noh},
  \bibinfo{person}{Long Ouyang}, \bibinfo{person}{Cullen O'Keefe},
  \bibinfo{person}{Jakub Pachocki}, \bibinfo{person}{Alex Paino},
  \bibinfo{person}{Joe Palermo}, \bibinfo{person}{Ashley Pantuliano},
  \bibinfo{person}{Giambattista Parascandolo}, \bibinfo{person}{Joel Parish},
  \bibinfo{person}{Emy Parparita}, \bibinfo{person}{Alex Passos},
  \bibinfo{person}{Mikhail Pavlov}, \bibinfo{person}{Andrew Peng},
  \bibinfo{person}{Adam Perelman}, \bibinfo{person}{Filipe de Avila
  Belbute~Peres}, \bibinfo{person}{Michael Petrov},
  \bibinfo{person}{Henrique~Ponde de Oliveira~Pinto},
  \bibinfo{person}{Michael}, \bibinfo{person}{Pokorny},
  \bibinfo{person}{Michelle Pokrass}, \bibinfo{person}{Vitchyr Pong},
  \bibinfo{person}{Tolly Powell}, \bibinfo{person}{Alethea Power},
  \bibinfo{person}{Boris Power}, \bibinfo{person}{Elizabeth Proehl},
  \bibinfo{person}{Raul Puri}, \bibinfo{person}{Alec Radford},
  \bibinfo{person}{Jack Rae}, \bibinfo{person}{Aditya Ramesh},
  \bibinfo{person}{Cameron Raymond}, \bibinfo{person}{Francis Real},
  \bibinfo{person}{Kendra Rimbach}, \bibinfo{person}{Carl Ross},
  \bibinfo{person}{Bob Rotsted}, \bibinfo{person}{Henri Roussez},
  \bibinfo{person}{Nick Ryder}, \bibinfo{person}{Mario Saltarelli},
  \bibinfo{person}{Ted Sanders}, \bibinfo{person}{Shibani Santurkar},
  \bibinfo{person}{Girish Sastry}, \bibinfo{person}{Heather Schmidt},
  \bibinfo{person}{David Schnurr}, \bibinfo{person}{John Schulman},
  \bibinfo{person}{Daniel Selsam}, \bibinfo{person}{Kyla Sheppard},
  \bibinfo{person}{Toki Sherbakov}, \bibinfo{person}{Jessica Shieh},
  \bibinfo{person}{Sarah Shoker}, \bibinfo{person}{Pranav Shyam},
  \bibinfo{person}{Szymon Sidor}, \bibinfo{person}{Eric Sigler},
  \bibinfo{person}{Maddie Simens}, \bibinfo{person}{Jordan Sitkin},
  \bibinfo{person}{Katarina Slama}, \bibinfo{person}{Ian Sohl},
  \bibinfo{person}{Benjamin Sokolowsky}, \bibinfo{person}{Yang Song},
  \bibinfo{person}{Natalie Staudacher}, \bibinfo{person}{Felipe~Petroski Such},
  \bibinfo{person}{Natalie Summers}, \bibinfo{person}{Ilya Sutskever},
  \bibinfo{person}{Jie Tang}, \bibinfo{person}{Nikolas Tezak},
  \bibinfo{person}{Madeleine Thompson}, \bibinfo{person}{Phil Tillet},
  \bibinfo{person}{Amin Tootoonchian}, \bibinfo{person}{Elizabeth Tseng},
  \bibinfo{person}{Preston Tuggle}, \bibinfo{person}{Nick Turley},
  \bibinfo{person}{Jerry Tworek}, \bibinfo{person}{Juan Felipe~Cerón Uribe},
  \bibinfo{person}{Andrea Vallone}, \bibinfo{person}{Arun Vijayvergiya},
  \bibinfo{person}{Chelsea Voss}, \bibinfo{person}{Carroll Wainwright},
  \bibinfo{person}{Justin~Jay Wang}, \bibinfo{person}{Alvin Wang},
  \bibinfo{person}{Ben Wang}, \bibinfo{person}{Jonathan Ward},
  \bibinfo{person}{Jason Wei}, \bibinfo{person}{CJ Weinmann},
  \bibinfo{person}{Akila Welihinda}, \bibinfo{person}{Peter Welinder},
  \bibinfo{person}{Jiayi Weng}, \bibinfo{person}{Lilian Weng},
  \bibinfo{person}{Matt Wiethoff}, \bibinfo{person}{Dave Willner},
  \bibinfo{person}{Clemens Winter}, \bibinfo{person}{Samuel Wolrich},
  \bibinfo{person}{Hannah Wong}, \bibinfo{person}{Lauren Workman},
  \bibinfo{person}{Sherwin Wu}, \bibinfo{person}{Jeff Wu},
  \bibinfo{person}{Michael Wu}, \bibinfo{person}{Kai Xiao},
  \bibinfo{person}{Tao Xu}, \bibinfo{person}{Sarah Yoo}, \bibinfo{person}{Kevin
  Yu}, \bibinfo{person}{Qiming Yuan}, \bibinfo{person}{Wojciech Zaremba},
  \bibinfo{person}{Rowan Zellers}, \bibinfo{person}{Chong Zhang},
  \bibinfo{person}{Marvin Zhang}, \bibinfo{person}{Shengjia Zhao},
  \bibinfo{person}{Tianhao Zheng}, \bibinfo{person}{Juntang Zhuang},
  \bibinfo{person}{William Zhuk}, {and} \bibinfo{person}{Barret Zoph}.}
  \bibinfo{year}{2023}\natexlab{}.
\newblock \bibinfo{title}{GPT-4 Technical Report}.
\newblock
\newblock
\showeprint[arxiv]{2303.08774}~[cs.CL]


\bibitem[OpenAI(2023)]%
        {2023GPT4VisionSC}
\bibfield{author}{\bibinfo{person}{OpenAI}.} \bibinfo{year}{2023}\natexlab{}.
\newblock \showarticletitle{GPT-4V(ision) System Card}.
\newblock
\urldef\tempurl%
\url{https://api.semanticscholar.org/CorpusID:263218031}
\showURL{%
\tempurl}


\bibitem[Ouyang et~al\mbox{.}(2022)]%
        {ouyang2022training}
\bibfield{author}{\bibinfo{person}{Long Ouyang}, \bibinfo{person}{Jeffrey Wu},
  \bibinfo{person}{Xu Jiang}, \bibinfo{person}{Diogo Almeida},
  \bibinfo{person}{Carroll Wainwright}, \bibinfo{person}{Pamela Mishkin},
  \bibinfo{person}{Chong Zhang}, \bibinfo{person}{Sandhini Agarwal},
  \bibinfo{person}{Katarina Slama}, \bibinfo{person}{Alex Ray},
  {et~al\mbox{.}}} \bibinfo{year}{2022}\natexlab{}.
\newblock \showarticletitle{Training language models to follow instructions
  with human feedback}.
\newblock \bibinfo{journal}{\emph{Advances in Neural Information Processing
  Systems}}  \bibinfo{volume}{35} (\bibinfo{year}{2022}),
  \bibinfo{pages}{27730--27744}.
\newblock


\bibitem[Pan et~al\mbox{.}(2023)]%
        {pan2023fact}
\bibfield{author}{\bibinfo{person}{Liangming Pan}, \bibinfo{person}{Xiaobao
  Wu}, \bibinfo{person}{Xinyuan Lu}, \bibinfo{person}{Anh~Tuan Luu},
  \bibinfo{person}{William~Yang Wang}, \bibinfo{person}{Min-Yen Kan}, {and}
  \bibinfo{person}{Preslav Nakov}.} \bibinfo{year}{2023}\natexlab{}.
\newblock \showarticletitle{Fact-Checking Complex Claims with Program-Guided
  Reasoning}.
\newblock \bibinfo{journal}{\emph{arXiv preprint arXiv:2305.12744}}
  (\bibinfo{year}{2023}).
\newblock


\bibitem[Pavlyshenko(2023)]%
        {pavlyshenko2023analysis}
\bibfield{author}{\bibinfo{person}{Bohdan~M Pavlyshenko}.}
  \bibinfo{year}{2023}\natexlab{}.
\newblock \showarticletitle{Analysis of disinformation and fake news detection
  using fine-tuned large language model}.
\newblock \bibinfo{journal}{\emph{arXiv preprint arXiv:2309.04704}}
  (\bibinfo{year}{2023}).
\newblock


\bibitem[Pelrine et~al\mbox{.}(2023)]%
        {pelrine2023towards}
\bibfield{author}{\bibinfo{person}{Kellin Pelrine}, \bibinfo{person}{Meilina
  Reksoprodjo}, \bibinfo{person}{Caleb Gupta}, \bibinfo{person}{Joel
  Christoph}, {and} \bibinfo{person}{Reihaneh Rabbany}.}
  \bibinfo{year}{2023}\natexlab{}.
\newblock \showarticletitle{Towards Reliable Misinformation Mitigation:
  Generalization, Uncertainty, and GPT-4}.
\newblock \bibinfo{journal}{\emph{arXiv preprint arXiv:2305.14928}}
  (\bibinfo{year}{2023}).
\newblock


\bibitem[Potthast et~al\mbox{.}(2017)]%
        {potthast2017stylometric}
\bibfield{author}{\bibinfo{person}{Martin Potthast}, \bibinfo{person}{Johannes
  Kiesel}, \bibinfo{person}{Kevin Reinartz}, \bibinfo{person}{Janek
  Bevendorff}, {and} \bibinfo{person}{Benno Stein}.}
  \bibinfo{year}{2017}\natexlab{}.
\newblock \showarticletitle{A stylometric inquiry into hyperpartisan and fake
  news}.
\newblock \bibinfo{journal}{\emph{arXiv preprint arXiv:1702.05638}}
  (\bibinfo{year}{2017}).
\newblock


\bibitem[Qian et~al\mbox{.}(2018)]%
        {qian2018neural}
\bibfield{author}{\bibinfo{person}{Feng Qian}, \bibinfo{person}{Chengyue Gong},
  \bibinfo{person}{Karishma Sharma}, {and} \bibinfo{person}{Yan Liu}.}
  \bibinfo{year}{2018}\natexlab{}.
\newblock \showarticletitle{Neural User Response Generator: Fake News Detection
  with Collective User Intelligence.}. In \bibinfo{booktitle}{\emph{IJCAI}},
  Vol.~\bibinfo{volume}{18}. \bibinfo{pages}{3834--3840}.
\newblock


\bibitem[Ramnath et~al\mbox{.}(2023)]%
        {ramnath2023tailoring}
\bibfield{author}{\bibinfo{person}{Sahana Ramnath}, \bibinfo{person}{Brihi
  Joshi}, \bibinfo{person}{Skyler Hallinan}, \bibinfo{person}{Ximing Lu},
  \bibinfo{person}{Liunian~Harold Li}, \bibinfo{person}{Aaron Chan},
  \bibinfo{person}{Jack Hessel}, \bibinfo{person}{Yejin Choi}, {and}
  \bibinfo{person}{Xiang Ren}.} \bibinfo{year}{2023}\natexlab{}.
\newblock \bibinfo{title}{Tailoring Self-Rationalizers with Multi-Reward
  Distillation}.
\newblock
\newblock
\showeprint[arxiv]{2311.02805}~[cs.CL]


\bibitem[Roy et~al\mbox{.}(2023)]%
        {roy2023probing}
\bibfield{author}{\bibinfo{person}{Sarthak Roy}, \bibinfo{person}{Ashish
  Harshavardhan}, \bibinfo{person}{Animesh Mukherjee}, {and}
  \bibinfo{person}{Punyajoy Saha}.} \bibinfo{year}{2023}\natexlab{}.
\newblock \showarticletitle{Probing LLMs for hate speech detection: strengths
  and vulnerabilities}.
\newblock \bibinfo{journal}{\emph{arXiv preprint arXiv:2310.12860}}
  (\bibinfo{year}{2023}).
\newblock


\bibitem[Rubin et~al\mbox{.}(2022)]%
        {DBLP:conf/naacl/RubinHB22}
\bibfield{author}{\bibinfo{person}{Ohad Rubin}, \bibinfo{person}{Jonathan
  Herzig}, {and} \bibinfo{person}{Jonathan Berant}.}
  \bibinfo{year}{2022}\natexlab{}.
\newblock \showarticletitle{Learning To Retrieve Prompts for In-Context
  Learning}. In \bibinfo{booktitle}{\emph{Proceedings of the 2022 Conference of
  the North American Chapter of the Association for Computational Linguistics:
  Human Language Technologies, {NAACL} 2022, Seattle, WA, United States, July
  10-15, 2022}}, \bibfield{editor}{\bibinfo{person}{Marine Carpuat},
  \bibinfo{person}{Marie{-}Catherine de~Marneffe}, {and}
  \bibinfo{person}{Iv{\'{a}}n Vladimir~Meza Ru{\'{\i}}z}} (Eds.).
  \bibinfo{publisher}{Association for Computational Linguistics},
  \bibinfo{pages}{2655--2671}.
\newblock
\urldef\tempurl%
\url{https://doi.org/10.18653/V1/2022.NAACL-MAIN.191}
\showDOI{\tempurl}


\bibitem[Sharma et~al\mbox{.}(2022)]%
        {DBLP:conf/ijcai/SharmaAADMFHSN022}
\bibfield{author}{\bibinfo{person}{Shivam Sharma}, \bibinfo{person}{Firoj
  Alam}, \bibinfo{person}{Md.~Shad Akhtar}, \bibinfo{person}{Dimitar Dimitrov},
  \bibinfo{person}{Giovanni Da~San Martino}, \bibinfo{person}{Hamed Firooz},
  \bibinfo{person}{Alon~Y. Halevy}, \bibinfo{person}{Fabrizio Silvestri},
  \bibinfo{person}{Preslav Nakov}, {and} \bibinfo{person}{Tanmoy Chakraborty}.}
  \bibinfo{year}{2022}\natexlab{}.
\newblock \showarticletitle{Detecting and Understanding Harmful Memes: {A}
  Survey}. In \bibinfo{booktitle}{\emph{Proceedings of the Thirty-First
  International Joint Conference on Artificial Intelligence, {IJCAI} 2022,
  Vienna, Austria, 23-29 July 2022}}, \bibfield{editor}{\bibinfo{person}{Luc~De
  Raedt}} (Ed.). \bibinfo{publisher}{ijcai.org}, \bibinfo{pages}{5597--5606}.
\newblock
\urldef\tempurl%
\url{https://doi.org/10.24963/IJCAI.2022/781}
\showDOI{\tempurl}


\bibitem[Shi et~al\mbox{.}(2023)]%
        {shi2023replug}
\bibfield{author}{\bibinfo{person}{Weijia Shi}, \bibinfo{person}{Sewon Min},
  \bibinfo{person}{Michihiro Yasunaga}, \bibinfo{person}{Minjoon Seo},
  \bibinfo{person}{Rich James}, \bibinfo{person}{Mike Lewis},
  \bibinfo{person}{Luke Zettlemoyer}, {and} \bibinfo{person}{Wen-tau Yih}.}
  \bibinfo{year}{2023}\natexlab{}.
\newblock \showarticletitle{Replug: Retrieval-augmented black-box language
  models}.
\newblock \bibinfo{journal}{\emph{arXiv preprint arXiv:2301.12652}}
  (\bibinfo{year}{2023}).
\newblock


\bibitem[Shu et~al\mbox{.}(2019)]%
        {shu2019defend}
\bibfield{author}{\bibinfo{person}{Kai Shu}, \bibinfo{person}{Limeng Cui},
  \bibinfo{person}{Suhang Wang}, \bibinfo{person}{Dongwon Lee}, {and}
  \bibinfo{person}{Huan Liu}.} \bibinfo{year}{2019}\natexlab{}.
\newblock \showarticletitle{defend: Explainable fake news detection}. In
  \bibinfo{booktitle}{\emph{Proceedings of the 25th ACM SIGKDD international
  conference on knowledge discovery \& data mining}}.
  \bibinfo{pages}{395--405}.
\newblock


\bibitem[Singhal et~al\mbox{.}(2022)]%
        {singhal2022leveraging}
\bibfield{author}{\bibinfo{person}{Shivangi Singhal}, \bibinfo{person}{Tanisha
  Pandey}, \bibinfo{person}{Saksham Mrig}, \bibinfo{person}{Rajiv~Ratn Shah},
  {and} \bibinfo{person}{Ponnurangam Kumaraguru}.}
  \bibinfo{year}{2022}\natexlab{}.
\newblock \showarticletitle{Leveraging Intra and Inter Modality Relationship
  for Multimodal Fake News Detection}. In \bibinfo{booktitle}{\emph{Companion
  Proceedings of the Web Conference 2022}}. \bibinfo{pages}{726--734}.
\newblock


\bibitem[Sun et~al\mbox{.}(2023)]%
        {sun2023inconsistent}
\bibfield{author}{\bibinfo{person}{Mengzhu Sun}, \bibinfo{person}{Xi Zhang},
  \bibinfo{person}{Jianqiang Ma}, \bibinfo{person}{Sihong Xie},
  \bibinfo{person}{Yazheng Liu}, {and} \bibinfo{person}{S~Yu Philip}.}
  \bibinfo{year}{2023}\natexlab{}.
\newblock \showarticletitle{Inconsistent matters: A knowledge-guided
  dual-consistency network for multi-modal rumor detection}.
\newblock \bibinfo{journal}{\emph{IEEE Transactions on Knowledge and Data
  Engineering}} (\bibinfo{year}{2023}).
\newblock


\bibitem[Team et~al\mbox{.}(2023)]%
        {team2023gemini}
\bibfield{author}{\bibinfo{person}{Gemini Team}, \bibinfo{person}{Rohan Anil},
  \bibinfo{person}{Sebastian Borgeaud}, \bibinfo{person}{Yonghui Wu},
  \bibinfo{person}{Jean-Baptiste Alayrac}, \bibinfo{person}{Jiahui Yu},
  \bibinfo{person}{Radu Soricut}, \bibinfo{person}{Johan Schalkwyk},
  \bibinfo{person}{Andrew~M Dai}, \bibinfo{person}{Anja Hauth},
  {et~al\mbox{.}}} \bibinfo{year}{2023}\natexlab{}.
\newblock \showarticletitle{Gemini: a family of highly capable multimodal
  models}.
\newblock \bibinfo{journal}{\emph{arXiv preprint arXiv:2312.11805}}
  (\bibinfo{year}{2023}).
\newblock


\bibitem[Touvron et~al\mbox{.}(2023)]%
        {touvron2023llama}
\bibfield{author}{\bibinfo{person}{Hugo Touvron}, \bibinfo{person}{Louis
  Martin}, \bibinfo{person}{Kevin Stone}, \bibinfo{person}{Peter Albert},
  \bibinfo{person}{Amjad Almahairi}, \bibinfo{person}{Yasmine Babaei},
  \bibinfo{person}{Nikolay Bashlykov}, \bibinfo{person}{Soumya Batra},
  \bibinfo{person}{Prajjwal Bhargava}, \bibinfo{person}{Shruti Bhosale},
  {et~al\mbox{.}}} \bibinfo{year}{2023}\natexlab{}.
\newblock \showarticletitle{Llama 2: Open foundation and fine-tuned chat
  models}.
\newblock \bibinfo{journal}{\emph{arXiv preprint arXiv:2307.09288}}
  (\bibinfo{year}{2023}).
\newblock


\bibitem[Wan et~al\mbox{.}(2024)]%
        {wan2024dell}
\bibfield{author}{\bibinfo{person}{Herun Wan}, \bibinfo{person}{Shangbin Feng},
  \bibinfo{person}{Zhaoxuan Tan}, \bibinfo{person}{Heng Wang},
  \bibinfo{person}{Yulia Tsvetkov}, {and} \bibinfo{person}{Minnan Luo}.}
  \bibinfo{year}{2024}\natexlab{}.
\newblock \showarticletitle{DELL: Generating Reactions and Explanations for
  LLM-Based Misinformation Detection}.
\newblock \bibinfo{journal}{\emph{arXiv preprint arXiv:2402.10426}}
  (\bibinfo{year}{2024}).
\newblock


\bibitem[Wang et~al\mbox{.}(2023)]%
        {wang2023cross}
\bibfield{author}{\bibinfo{person}{Longzheng Wang}, \bibinfo{person}{Chuang
  Zhang}, \bibinfo{person}{Hongbo Xu}, \bibinfo{person}{Yongxiu Xu},
  \bibinfo{person}{Xiaohan Xu}, {and} \bibinfo{person}{Siqi Wang}.}
  \bibinfo{year}{2023}\natexlab{}.
\newblock \showarticletitle{Cross-modal contrastive learning for multimodal
  fake news detection}. In \bibinfo{booktitle}{\emph{Proceedings of the 31st
  ACM International Conference on Multimedia}}. \bibinfo{pages}{5696--5704}.
\newblock


\bibitem[Wang et~al\mbox{.}(2022a)]%
        {wang2022self}
\bibfield{author}{\bibinfo{person}{Yizhong Wang}, \bibinfo{person}{Yeganeh
  Kordi}, \bibinfo{person}{Swaroop Mishra}, \bibinfo{person}{Alisa Liu},
  \bibinfo{person}{Noah~A Smith}, \bibinfo{person}{Daniel Khashabi}, {and}
  \bibinfo{person}{Hannaneh Hajishirzi}.} \bibinfo{year}{2022}\natexlab{a}.
\newblock \showarticletitle{Self-instruct: Aligning language models with
  self-generated instructions}.
\newblock \bibinfo{journal}{\emph{arXiv preprint arXiv:2212.10560}}
  (\bibinfo{year}{2022}).
\newblock


\bibitem[Wang et~al\mbox{.}(2018)]%
        {wang2018eann}
\bibfield{author}{\bibinfo{person}{Yaqing Wang}, \bibinfo{person}{Fenglong Ma},
  \bibinfo{person}{Zhiwei Jin}, \bibinfo{person}{Ye Yuan},
  \bibinfo{person}{Guangxu Xun}, \bibinfo{person}{Kishlay Jha},
  \bibinfo{person}{Lu Su}, {and} \bibinfo{person}{Jing Gao}.}
  \bibinfo{year}{2018}\natexlab{}.
\newblock \showarticletitle{Eann: Event adversarial neural networks for
  multi-modal fake news detection}. In \bibinfo{booktitle}{\emph{Proceedings of
  the 24th acm sigkdd international conference on knowledge discovery \& data
  mining}}. \bibinfo{pages}{849--857}.
\newblock


\bibitem[Wang et~al\mbox{.}(2022b)]%
        {wang2022benchmarking}
\bibfield{author}{\bibinfo{person}{Yizhong Wang}, \bibinfo{person}{Swaroop
  Mishra}, \bibinfo{person}{Pegah Alipoormolabashi}, \bibinfo{person}{Yeganeh
  Kordi}, \bibinfo{person}{Amirreza Mirzaei}, \bibinfo{person}{Anjana
  Arunkumar}, \bibinfo{person}{Arjun Ashok}, \bibinfo{person}{Arut~Selvan
  Dhanasekaran}, \bibinfo{person}{Atharva Naik}, \bibinfo{person}{David Stap},
  {et~al\mbox{.}}} \bibinfo{year}{2022}\natexlab{b}.
\newblock \showarticletitle{Benchmarking generalization via in-context
  instructions on 1,600+ language tasks}.
\newblock \bibinfo{journal}{\emph{arXiv preprint arXiv:2204.07705}}
  \bibinfo{volume}{2} (\bibinfo{year}{2022}).
\newblock


\bibitem[Wei et~al\mbox{.}(2021)]%
        {wei2021finetuned}
\bibfield{author}{\bibinfo{person}{Jason Wei}, \bibinfo{person}{Maarten Bosma},
  \bibinfo{person}{Vincent Zhao}, \bibinfo{person}{Kelvin Guu},
  \bibinfo{person}{Adams~Wei Yu}, \bibinfo{person}{Brian Lester},
  \bibinfo{person}{Nan Du}, \bibinfo{person}{Andrew~M Dai}, {and}
  \bibinfo{person}{Quoc~V Le}.} \bibinfo{year}{2021}\natexlab{}.
\newblock \showarticletitle{Finetuned Language Models are Zero-Shot Learners}.
  In \bibinfo{booktitle}{\emph{International Conference on Learning
  Representations}}.
\newblock


\bibitem[Wu et~al\mbox{.}(2021)]%
        {wu2021multimodal}
\bibfield{author}{\bibinfo{person}{Yang Wu}, \bibinfo{person}{Pengwei Zhan},
  \bibinfo{person}{Yunjian Zhang}, \bibinfo{person}{Liming Wang}, {and}
  \bibinfo{person}{Zhen Xu}.} \bibinfo{year}{2021}\natexlab{}.
\newblock \showarticletitle{Multimodal fusion with co-attention networks for
  fake news detection}. In \bibinfo{booktitle}{\emph{Findings of the
  Association for Computational Linguistics: ACL-IJCNLP 2021}}.
  \bibinfo{pages}{2560--2569}.
\newblock


\bibitem[Xu et~al\mbox{.}(2023)]%
        {xu2023wizardlm}
\bibfield{author}{\bibinfo{person}{Can Xu}, \bibinfo{person}{Qingfeng Sun},
  \bibinfo{person}{Kai Zheng}, \bibinfo{person}{Xiubo Geng},
  \bibinfo{person}{Pu Zhao}, \bibinfo{person}{Jiazhan Feng},
  \bibinfo{person}{Chongyang Tao}, {and} \bibinfo{person}{Daxin Jiang}.}
  \bibinfo{year}{2023}\natexlab{}.
\newblock \showarticletitle{Wizardlm: Empowering large language models to
  follow complex instructions}.
\newblock \bibinfo{journal}{\emph{arXiv preprint arXiv:2304.12244}}
  (\bibinfo{year}{2023}).
\newblock


\bibitem[Xu et~al\mbox{.}(2024)]%
        {xu2024survey}
\bibfield{author}{\bibinfo{person}{Xiaohan Xu}, \bibinfo{person}{Ming Li},
  \bibinfo{person}{Chongyang Tao}, \bibinfo{person}{Tao Shen},
  \bibinfo{person}{Reynold Cheng}, \bibinfo{person}{Jinyang Li},
  \bibinfo{person}{Can Xu}, \bibinfo{person}{Dacheng Tao}, {and}
  \bibinfo{person}{Tianyi Zhou}.} \bibinfo{year}{2024}\natexlab{}.
\newblock \showarticletitle{A survey on knowledge distillation of large
  language models}.
\newblock \bibinfo{journal}{\emph{arXiv preprint arXiv:2402.13116}}
  (\bibinfo{year}{2024}).
\newblock


\bibitem[Yang et~al\mbox{.}(2023)]%
        {yang2023gpt4tools}
\bibfield{author}{\bibinfo{person}{Rui Yang}, \bibinfo{person}{Lin Song},
  \bibinfo{person}{Yanwei Li}, \bibinfo{person}{Sijie Zhao},
  \bibinfo{person}{Yixiao Ge}, \bibinfo{person}{Xiu Li}, {and}
  \bibinfo{person}{Ying Shan}.} \bibinfo{year}{2023}\natexlab{}.
\newblock \bibinfo{title}{GPT4Tools: Teaching Large Language Model to Use Tools
  via Self-instruction}.
\newblock
\newblock
\showeprint[arxiv]{2305.18752}~[cs.CV]


\bibitem[Yang et~al\mbox{.}(2022)]%
        {DBLP:conf/coling/00050CLLC22}
\bibfield{author}{\bibinfo{person}{Zhiwei Yang}, \bibinfo{person}{Jing Ma},
  \bibinfo{person}{Hechang Chen}, \bibinfo{person}{Hongzhan Lin},
  \bibinfo{person}{Ziyang Luo}, {and} \bibinfo{person}{Yi Chang}.}
  \bibinfo{year}{2022}\natexlab{}.
\newblock \showarticletitle{A Coarse-to-fine Cascaded Evidence-Distillation
  Neural Network for Explainable Fake News Detection}. In
  \bibinfo{booktitle}{\emph{Proceedings of the 29th International Conference on
  Computational Linguistics, {COLING} 2022, Gyeongju, Republic of Korea,
  October 12-17, 2022}}, \bibfield{editor}{\bibinfo{person}{Nicoletta
  Calzolari}, \bibinfo{person}{Chu{-}Ren Huang}, \bibinfo{person}{Hansaem Kim},
  \bibinfo{person}{James Pustejovsky}, \bibinfo{person}{Leo Wanner},
  \bibinfo{person}{Key{-}Sun Choi}, \bibinfo{person}{Pum{-}Mo Ryu},
  \bibinfo{person}{Hsin{-}Hsi Chen}, \bibinfo{person}{Lucia Donatelli},
  \bibinfo{person}{Heng Ji}, \bibinfo{person}{Sadao Kurohashi},
  \bibinfo{person}{Patrizia Paggio}, \bibinfo{person}{Nianwen Xue},
  \bibinfo{person}{Seokhwan Kim}, \bibinfo{person}{Younggyun Hahm},
  \bibinfo{person}{Zhong He}, \bibinfo{person}{Tony~Kyungil Lee},
  \bibinfo{person}{Enrico Santus}, \bibinfo{person}{Francis Bond}, {and}
  \bibinfo{person}{Seung{-}Hoon Na}} (Eds.). \bibinfo{publisher}{International
  Committee on Computational Linguistics}, \bibinfo{pages}{2608--2621}.
\newblock
\urldef\tempurl%
\url{https://aclanthology.org/2022.coling-1.230}
\showURL{%
\tempurl}


\bibitem[Yu et~al\mbox{.}(2017)]%
        {yu2017convolutional}
\bibfield{author}{\bibinfo{person}{Feng Yu}, \bibinfo{person}{Qiang Liu},
  \bibinfo{person}{Shu Wu}, \bibinfo{person}{Liang Wang},
  \bibinfo{person}{Tieniu Tan}, {et~al\mbox{.}}}
  \bibinfo{year}{2017}\natexlab{}.
\newblock \showarticletitle{A Convolutional Approach for Misinformation
  Identification.}. In \bibinfo{booktitle}{\emph{IJCAI}}.
  \bibinfo{pages}{3901--3907}.
\newblock


\bibitem[Zhang and Gao(2023)]%
        {zhang2023towards}
\bibfield{author}{\bibinfo{person}{Xuan Zhang} {and} \bibinfo{person}{Wei
  Gao}.} \bibinfo{year}{2023}\natexlab{}.
\newblock \showarticletitle{Towards llm-based fact verification on news claims
  with a hierarchical step-by-step prompting method}.
\newblock \bibinfo{journal}{\emph{arXiv preprint arXiv:2310.00305}}
  (\bibinfo{year}{2023}).
\newblock


\bibitem[Zubiaga et~al\mbox{.}(2018)]%
        {zubiaga2018detection}
\bibfield{author}{\bibinfo{person}{Arkaitz Zubiaga}, \bibinfo{person}{Ahmet
  Aker}, \bibinfo{person}{Kalina Bontcheva}, \bibinfo{person}{Maria Liakata},
  {and} \bibinfo{person}{Rob Procter}.} \bibinfo{year}{2018}\natexlab{}.
\newblock \showarticletitle{Detection and resolution of rumours in social
  media: A survey}.
\newblock \bibinfo{journal}{\emph{ACM Computing Surveys (CSUR)}}
  \bibinfo{volume}{51}, \bibinfo{number}{2} (\bibinfo{year}{2018}),
  \bibinfo{pages}{1--36}.
\newblock


\end{thebibliography}










\end{document}